\documentclass[10pt,twocolumn,letterpaper]{article}

\usepackage[pagenumbers]{iccv} % To force page numbers, e.g. for an arXiv version

\definecolor{iccvblue}{rgb}{0.21,0.49,0.74}
\usepackage[pagebackref,breaklinks,colorlinks,allcolors=iccvblue]{hyperref}

\usepackage{xcolor}
\usepackage{colortbl}
\usepackage{tcolorbox}
\usepackage{amsmath}
\usepackage{kotex}
\usepackage{graphicx}
\usepackage{multirow}
\usepackage{caption}
\usepackage{subcaption}
\usepackage{booktabs}
\usepackage{comment}
\usepackage{algorithm}
\usepackage{algpseudocode}
\usepackage{bm}
\usepackage{afterpage}

\newcommand\blfootnote[1]{%
  \begingroup
  \renewcommand\thefootnote{}\footnote{#1}%
  \addtocounter{footnote}{-1}%
  \endgroup
}
\tcbuselibrary{listingsutf8} % Enable listings support in tcolorbox
\newtcblisting{pythoncode}{
    colback=gray!10, % Background color
    colframe=black, % Frame color
    listing only, % Code-only mode
    listing options={
        basicstyle=\ttfamily\footnotesize, % Font style and size
        breaklines=true, % Enable line breaking
        numbers=left, % Add line numbers
        numberstyle=\tiny\color{gray}, % Line number style
        keywordstyle=\color{blue}, % Keywords in blue
        commentstyle=\color{green!50!black}, % Comments in green
        stringstyle=\color{red}, % Strings in red
        language=Python % Language setting
    }
}

\def\paperID{} % *** Enter the Paper ID here
\def\confName{ICCV}
\def\confYear{2025}

\title{Style-Friendly SNR Sampler for Style-Driven Generation}

\author{Jooyoung Choi$^{1,*}$~~~~~Chaehun Shin$^{1,*}$~~~~~Yeongtak Oh$^1$~~~~~Heeseung Kim$^1$ \\ Jungbeom Lee$^2$~~~~~Sungroh Yoon$^{1,3,\dagger}$\\
$^1$Data Science and AI Laboratory, ECE, Seoul National University\\
$^2$Amazon\\
$^3$AIIS, ASRI, INMC, ISRC, and Interdisciplinary Program in AI, Seoul National University\\
{\tt\small \{jy\_choi,chaehuny,dualism9306,gmltmd789\}@snu.ac.kr,jungbeol@amazon.com,sryoon@snu.ac.kr} \\
{\small \url{https://stylefriendly.github.io}}
}

\begin{document}

\twocolumn[{
\renewcommand\twocolumn[1][]{#1}
\maketitle

\begin{center} 
\centering
    \centering
    \includegraphics[width=\textwidth]{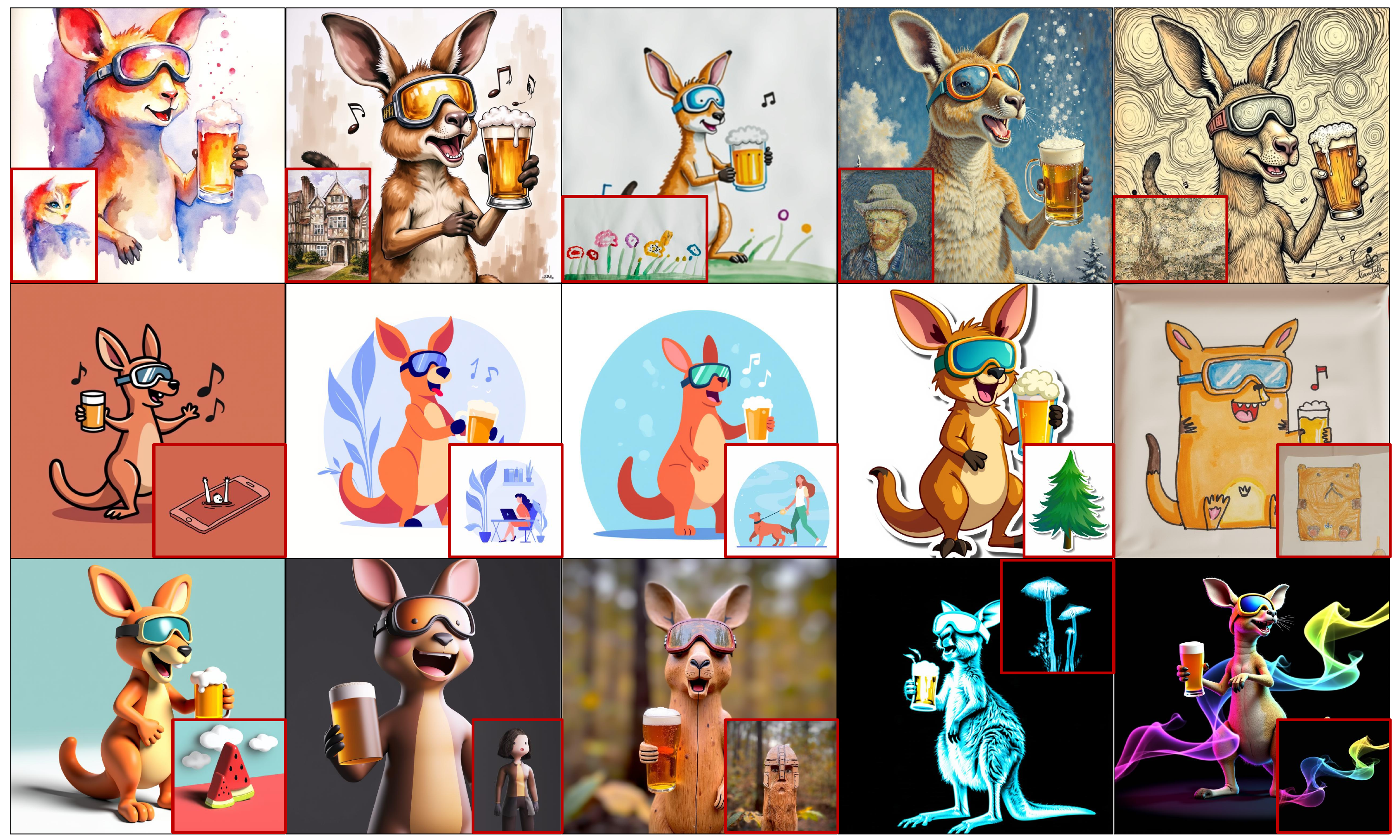}
    \captionof{figure}{Fine-tuning text-to-image diffusion models on the \textit{style-friendly} noise levels enables learning novel styles from reference images and text prompts. We present `A kangaroo holding a beer, wearing ski goggles and passionately singing silly songs' in various styles including watercolor painting, flat illustration, and 3d rendering styles. References are shown in the red insert box.}
    \label{fig:teaser}
\end{center}
}]
\blfootnote{$\dagger$ Correspondence to: Sungroh Yoon (sryoon@snu.ac.kr)}
\blfootnote{$*$ Both authors contributed equally to this work}

\begin{abstract}
Recent text-to-image diffusion models generate high-quality images but struggle to learn new, personalized styles, which limits the creation of unique style templates. In style-driven generation, users typically supply reference images exemplifying the desired style, together with text prompts that specify desired stylistic attributes. Previous approaches popularly rely on fine-tuning, yet it often blindly utilizes objectives and noise level distributions from pre-training without adaptation. We discover that stylistic features predominantly emerge at higher noise levels, leading current fine-tuning methods to exhibit suboptimal style alignment. We propose the Style-friendly SNR sampler, which aggressively shifts the signal-to-noise ratio (SNR) distribution toward higher noise levels during fine-tuning to focus on noise levels where stylistic features emerge. This enhances models' ability to capture novel styles indicated by reference images and text prompts. We demonstrate improved generation of novel styles that cannot be adequately described solely with a text prompt, enabling the creation of new style templates for personalized content creation.
\end{abstract}  
\section{Introduction}
\label{sec:intro}

Recently, large-scale text-to-image diffusion models~\cite{ldm,sd3,sdxl,stablecascade,imagen,flux1-dev,ediff} have achieved remarkable progress in visual content creation. 
In particular, open-weights such as Stable Diffusion series~\cite{ldm,sd3} and FLUX~\cite{flux1-dev} have been among the most notable for their photorealistic image quality and language understanding capabilities.
Behind this strong performance lies the advancement of the diffusion framework that encompasses the principles of score-based models~\cite{sde} and flow matching~\cite{flowmatching, rectifiedflow}, diffusion formulations~\cite{sde,flowmatching,rectifiedflow,ddpm}, loss weighting~\cite{p2,understanding}, noise level scheduling~\cite{simple,edm}, and architectural improvements~\cite{dit,sd3,edm2}.
These advancements have predominantly focused on generating high-quality images with respect to object-centric benchmarks~\cite{t2icompbench,geneval} and metrics~\cite{fid,kynkaanniemirole}.

Motivated by the success of text-to-image models, there is a growing need for style-driven generation~\cite{styledrop,stylealign,rb-modulation,ipadapter,dco}, where the generated samples capture styles desired by individual users or artists. Here, ``style" encompasses various elements such as color schemes, layouts, illumination, and brushwork~\cite{gatys2016image,styledrop,li2017universal,quilting}, all contributing to the unique nuances of an image, and is typically specified or clarified using text prompts. However, relying solely on prompt engineering has its limitations in reflecting unique styles, especially those not present in the pre-training data.

To achieve more effective style-driven generation, previous research has explored methods such as fine-tuning~\cite{dreambooth,styledrop,dco}, image variation~\cite{ipadapter} and editing~\cite{stylealign}. These approaches commonly use style reference images and their accompanying text prompts to guide the generation process. 
However, they \textit{blindly} apply the same objective functions and noise level distributions used for pre-training—originally optimized for object-centric benchmarks~\cite{geneval,t2icompbench}—without
acknowledging that style images emphasize different visual factors such as color schemes, layout, and brushstrokes, rather than simply depicting specific objects.
Consequently, even with numerous style-driven generation studies~\cite{styledrop,dco,ipadapter,rb-modulation,stylealign}, we observe prevalent failure cases when fine-tuning diffusion models on style references. Addressing these failures sometimes requires two-stage training with heavy human feedback~\cite{styledrop}.

In this paper, we address these limitations by introducing the \textit{Style-friendly SNR sampler}, a method that ensures capturing novel styles during fine-tuning. Our approach is motivated by two key observations: 1) diffusion models~\cite{sd3,flux1-dev} struggle to learn new styles, and 2) styles emerge at higher noise levels. These observations reveal that standard fine-tuning procedures often devote a large portion of computation toward lower noise levels, which are less relevant for learning novel styles.
Building upon these observations, %we propose adjusting the noise level sampling in the diffusion model's objective function during fine-tuning. 
we propose reallocating fine-tuning computation on higher noise levels, where essential style attributes emerge.

Our method, focusing on the reallocation of noise level sampling during fine-tuning, enhances the fidelity of style-driven generation, capturing the novel styles indicated by text prompts while preserving the capability of depicting the desired content. Moreover, we unveil the key components that explain why these models excel at learning object-centric concepts but struggle with styles, providing deeper insights into the diffusion process for style-driven generation. Ultimately, our approach facilitates the creation of style templates from reference images, which can be easily shared and utilized to practitioners for content creation, expanding the capabilities of diffusion models.

\section{Training Diffusion Models}
\begin{figure}
    \centering
    \includegraphics[width=\linewidth]{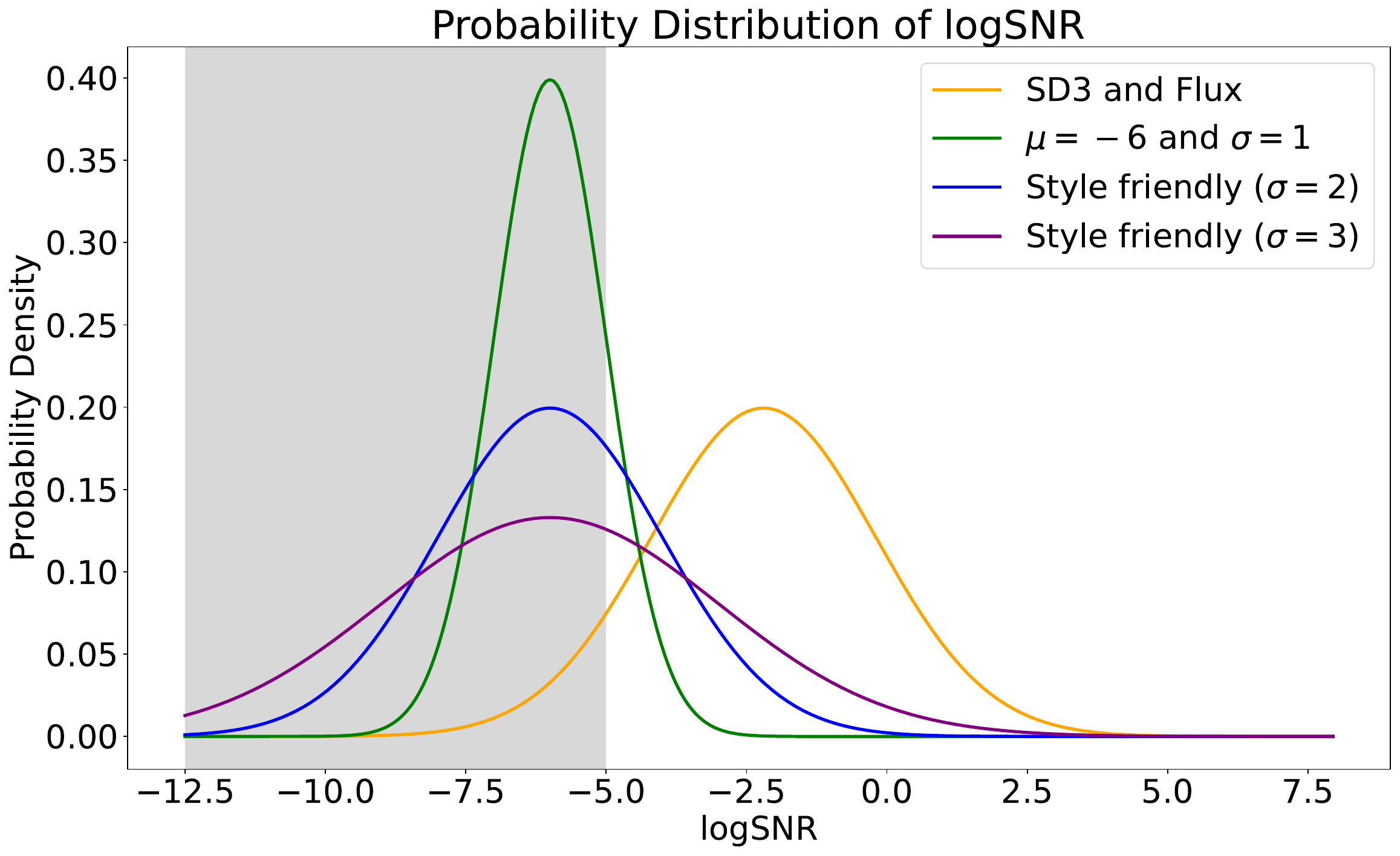}
  \caption{\textbf{Probability distribution of Log-SNR.} We bias the distribution towards the shaded region where style features emerge.}
  \label{fig:prob_logsnr}
\end{figure}

\subsection{Diffusion Process and SNR Formulation}

Various diffusion models~\cite{sohl2015deep,ddpm,sde,flowmatching,rectifiedflow} are based on the forward process that progressively degrades data $x_{0}$ into pure noise $x_{1}$ as time $t$ progresses from $0$ to $1$, following the unified formulation below:
\begin{equation}
  x_t=\alpha_tx_0+\sigma_t\epsilon,
  \label{eq:forward}
\end{equation}
where $\alpha_t$ and $\sigma_t$ are predefined noise schedules, and $\epsilon \sim \mathcal{N}(0,I)$ represents standard Gaussian noise.

%In discrete-time DDPM framework including Stable Diffusion XL (SDXL), the forward process involves a finite set of timesteps \( t \in \{0,1,\dots,T\} \), and set the $\alpha_t^2+\sigma_t^2=1$, $\alpha_t =\sqrt{\prod_{\tau=1}^{t} \bigl(1-\beta_\tau\bigr)}$ with linear schedule $\beta_t$.
Recent state-of-the-art flow matching frameworks, such as Stable Diffusion 3 (SD3)~\cite{sd3} and FLUX~\cite{flux1-dev}, utilize the noise schedule from rectified flow~\cite{flowmatching,rectifiedflow}, where $\alpha_t=1-t$ and $\sigma_t=t$, with $t$ varying continuously in the range $[0, 1]$. This choice is effective due to straight diffusion trajectories. 

Instead of parameterizing the diffusion process using the timestep $t$, Kingma et al.~\cite{vdm,understanding} characterize the noise level using the log signal-to-noise ratio (log-SNR), which offers a more intuitive measure of the noise at each step:
\begin{equation}
\lambda_t = \log\left(\frac{\alpha_t^2}{\sigma_t^2}\right).
\label{eq:logsnr}
\end{equation}
In flow matching framework, the log-SNR is simplified as $\lambda_t=2\log\left(\frac{1-t}{t}\right)$ using timestep $t$.
%In our work, we demonstrate that adjusting the timestep distribution based on $\lambda_t$ facilitates learning styles more effectively than using $t$.

\subsection{Diffusion Training and SNR Sampling}

Recent flow matching frameworks predict the velocity field $v_{\theta}(x_{t}, t)$ by minimizing the following training objective:
\begin{equation}
    \label{eq:fm_loss}
    \mathcal{L}_{\text{FM}}(x_0) 
    \;=\; 
    \mathbb{E}_{t \sim p(t)}
    \Big[
      \big\|\epsilon - x_0 - v_{\theta}(x_t, t)\big\|^2
    \Big],
\end{equation}
where $p(t)$ is the timestep sampling distribution, $\epsilon - x_0$ is the target velocity derived from the forward process in \cref{eq:forward}.
The representative text-to-image models utilizing flow matching formulation, such as SD3~\cite{sd3} and FLUX~\cite{flux1-dev}, introduce a SNR sampler for training.
This sampler samples the \textit{logit} of $t$, defined as $\log\left(\frac{t}{1-t}\right)$ from $\mathcal{N}(\mu, \sigma^2)$, where the parameters $\mu$ and $\sigma$ are chosen as $0$ and $1$ to optimize CLIP~\cite{clip} and FID scores~\cite{fid} on COCO-2014 validation set~\cite{coco}. 

In addition, they propose shifting timestep $t$ to $t_{\text{new}}$ by $k$ for high resolution training:
\begin{equation} 
t_{\text{new}} = \frac{k t}{1 + (k - 1) t},
\label{eq:t_shift}
\end{equation}
which is equivalent to shifting $\lambda_t$ by $-2\log k$ as follows:
\begin{equation} 
\lambda_{t_{\text{new}}} = 2 \log\left(\frac{1 - t_{\text{new}}}{t_{\text{new}}}\right) = \lambda_t - 2 \log k,
\label{eq:lambda_shift}
\end{equation}
where $k$ is defined as $3$.
Following the above formulation, resulting log-SNR sampling distribution $p(\lambda_t)$ in training time is represented as $\mathcal{N}(-2\log3, 2^2)$, as visualized by the yellow curve in \cref{fig:prob_logsnr}. This curve demonstrates that pre-training SD3 and FLUX focus on particular noise levels.

\section{Method}
\label{sec:method}

\subsection{Observations}
\label{sec:observation}

\begin{figure}[t]
  \centering
   \includegraphics[width=1.0\linewidth]{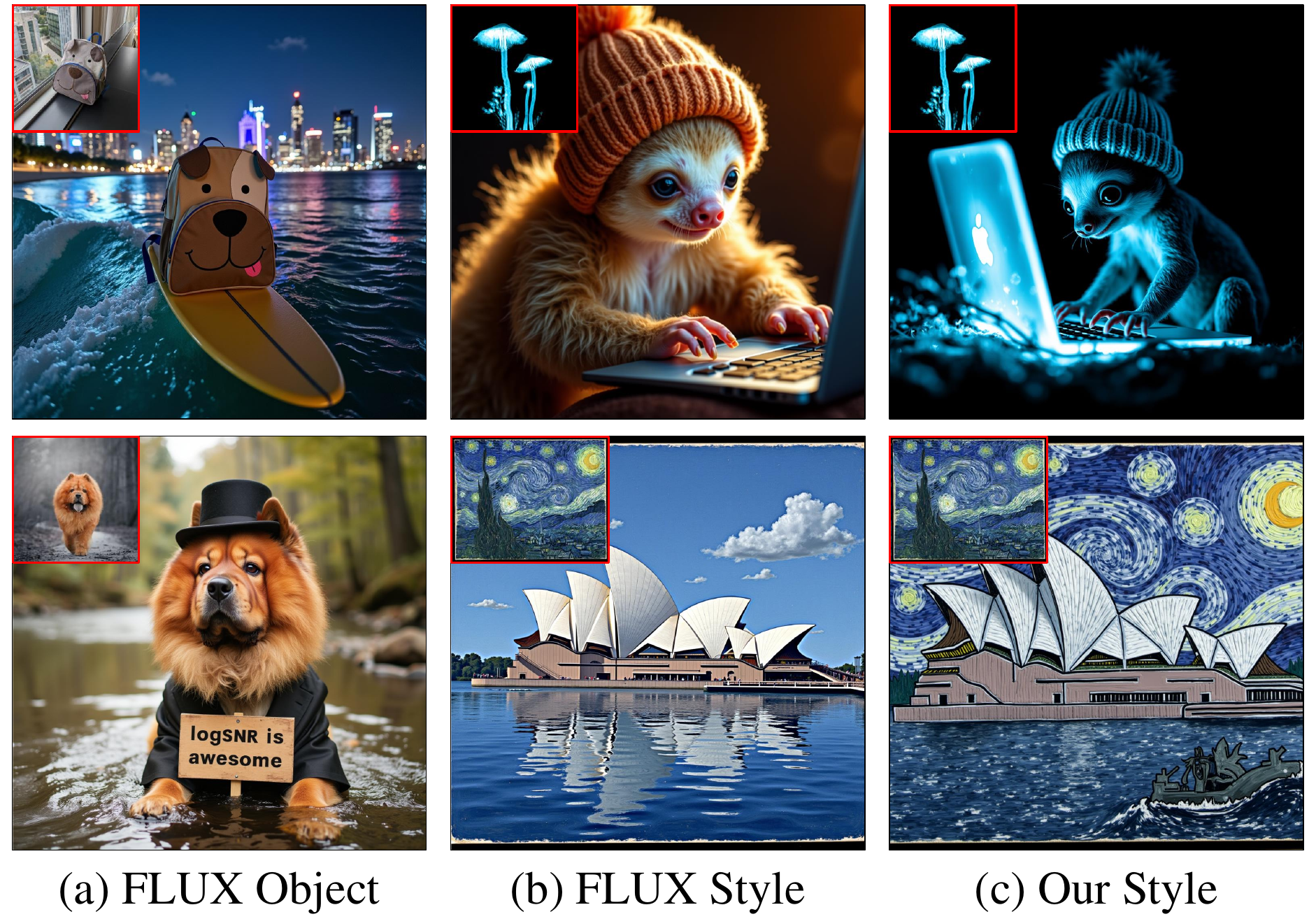}
   \caption{\textbf{Fine-tuning capability.} (a) While FLUX succeeds in learning objects, (b) it struggles to capture styles, demonstrating that learning novel objects and styles requires distinct strategies. (c) We enable FLUX to learn styles. References are shown in the red insert box.}
   \label{fig:dreambooth}
\end{figure}

\begin{figure*}[t]
  \centering
   \includegraphics[width=1.0\linewidth]{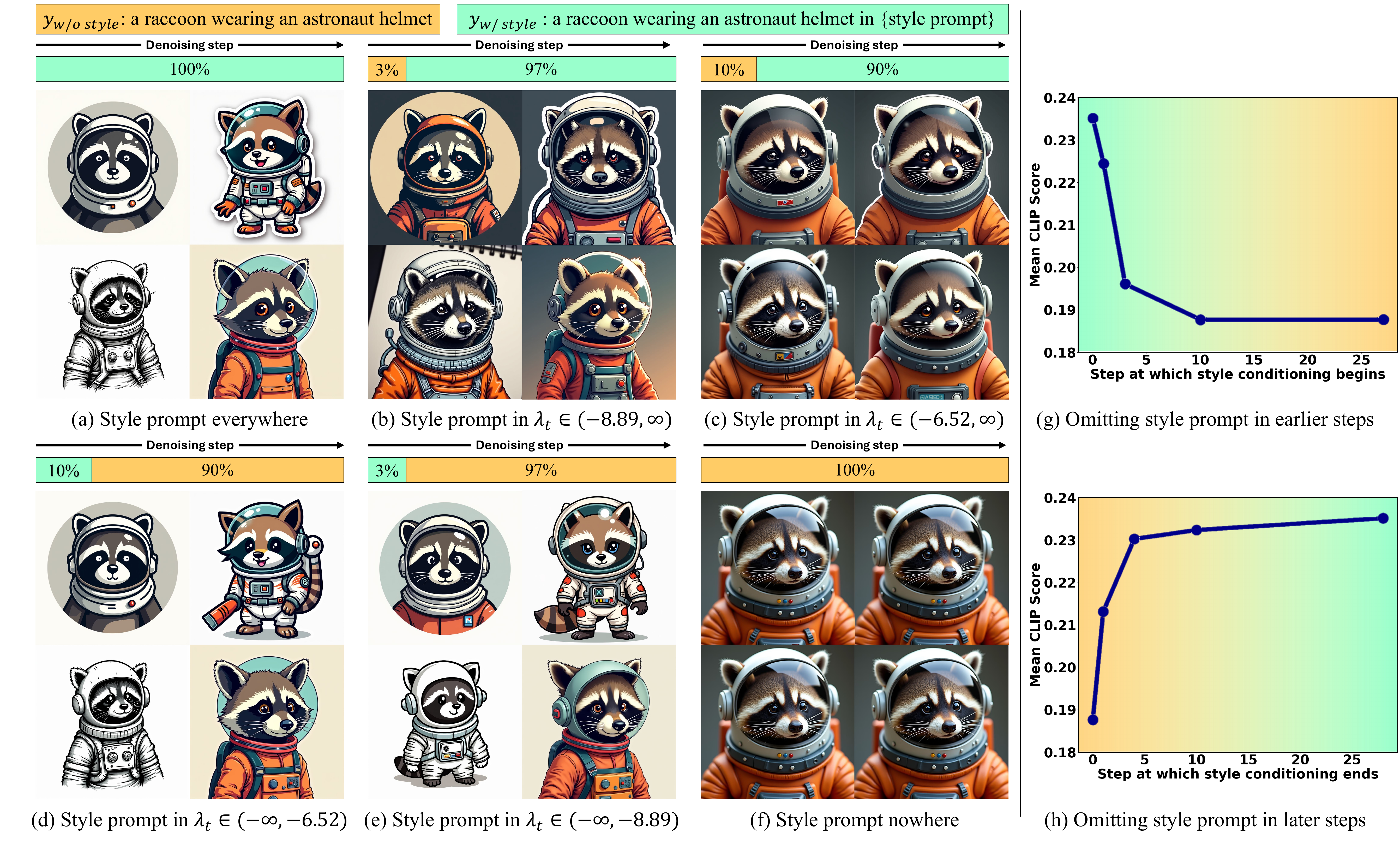}
   \caption{\textbf{Prompt switching during generation.} $\lambda_t$ indicates log-SNR. The bar graphs above each image represent the denoising steps, illustrating when each prompt is applied and at what point the prompt switch occurs. The style prompts are `minimalist flat round logo', `sticker', `detailed pen and ink drawing', and `cartoon'. Styles emerge in the initial 10\% of denoising steps; therefore, (c) and (f) fail to capture target styles. In contrast, omitting style prompts in later steps (d,e) still preserves styles well, similar to the fully styled baseline (a). (g) and (h) quantify these observations, showing the average CLIP similarity across 5 prompts and 5 styles when omitting (g) or including (h) the style prompt in earlier steps. Here, we use FLUX with 28 inference steps.}
   \label{fig:switch}
\end{figure*}

\paragraph{Diffusion Models Struggle to Capture Styles.}

We begin by examining the fine-tuning capabilities of current text-to-image diffusion models, which have primarily been studied for object-driven generation~\cite{dreambooth} from reference images. Fine-tuning is typically guided by accompanying text prompts that specify the desired objects or styles. As shown in \cref{fig:dreambooth}a, these methods often excel at producing high-quality object-driven images. However, we observe poorer results when they are applied to replicate distinctive color schemes, illumination, or brushwork—particularly when using the same SNR sampler from pre-training~\cite{sd3}.
%In \cref{fig:dreambooth}b, the ``glowing" example only causes the sloth's fur to glow but does not reflect the dark background and blue lighting of the reference. Similarly, for the Van Gogh oil painting reference, while it captures the blue color tone, it fails to reproduce Van Gogh's distinctive brushstrokes. 
In \cref{fig:dreambooth}b, when provided with a ``glowing" style reference, the FLUX model only applies glowing effects to specific object details (such as the fur of a sloth), neglecting broader stylistic elements like the dark background and blue lighting of the original reference. Similarly, with a Van Gogh oil painting style reference, the model manages to replicate the blue color tone but fails to accurately reproduce distinctive brushstroke characteristics.

\paragraph{Styles Emerge at Higher Noise Levels.}
To better understand why diffusion models struggle to capture new styles, we investigate at which noise levels stylistic features emerge during generation using a pre-trained FLUX model~\cite{flux1-dev}. Specifically, we switch from a prompt without style descriptions ($y_{\text{w/o style}}$) to one including style descriptions ($y_{\text{w/ style}}$) at different points in the denoising process. Omitting the style prompt in the initial \textbf{10\%} of generation steps significantly reduces style alignment (\cref{fig:switch}c), as quantitatively confirmed by the low CLIP similarity scores when initial denoising steps omit style information (\cref{fig:switch}g). Conversely, omitting the style prompt at later denoising steps minimally affects style alignment (\cref{fig:switch}d,e), which is also supported quantitatively in \cref{fig:switch}h. These results demonstrate that styles are predominantly determined at early denoising steps, corresponding to intervals with higher noise levels (low log-SNR $\lambda_t$ values).

\subsection{Style-Friendly SNR Sampler}

Our earlier observations show that styles primarily emerge during early denoising steps, characterized by higher noise levels (lower log-SNR values). However, existing fine-tuning methods use the SNR sampler from pre-training, optimized mainly for object-centric benchmarks~\cite{t2icompbench,geneval}, as indicated by the yellow curve in \cref{fig:prob_logsnr}. Consequently, standard fine-tuning procedures place insufficient emphasis on noise levels crucial for capturing styles, failing to achieve alignment with reference styles.

\begin{comment}
Our observations indicate the key motivation for style learning.
While styles emerge in the early steps of the denoising process, the current fine-tuning process utilizes an SNR sampler from pre-training which prioritizes object-centric generation~\cite{t2icompbench,geneval} as shown in the green line of \cref{fig:prob_logsnr}.
This SNR sampler places greater emphasis on intermediate steps to capture fine details of objects better, yet it does not sufficiently focus on the noise levels where styles emerge.
As a result, despite excelling in object-driven generation, the current fine-tuning struggles to fully capture and represent target styles in style-driven generation.
\end{comment}

Building upon this motivation, we propose to fine-tune diffusion models by biasing the noise level distribution towards higher noise levels (lower log-SNR $\lambda_t$ values) where stylistic features emerge. Specifically, we sample log-SNR from a normal distribution:
\begin{equation}
    \lambda_t \sim \mathcal{N}(\mu_{\text{low}},\sigma_{\text{large}}^2),
\end{equation}
with a lowered mean $\mu_{\text{low}}$, thereby focusing the training on higher noise levels essential for style learning. 
In addition, we choose sufficiently large $\sigma_{\text{large}}$ to cover the wide range of log-SNR values critical for style learning (shaded region in \cref{fig:prob_logsnr}). 
Finally, to map the sampled $\lambda_t$ back to the timestep domain, we compute $t \;=\; 1/(\,1 + \exp(\lambda_t/2)).$

While timestep shifting in \cite{sd3} weakly biases the noise level distribution as in \cref{eq:lambda_shift}, our Style-friendly SNR sampler aggressively targets the high-noise regions critical for capturing style. Setting the mean to $\mu_{\text{low}} = -6$ and $\sigma_{\text{large}} \ge 2$ targets the shaded region of \cref{fig:prob_logsnr}, substantially improving style alignment across a variety of reference styles.

\subsection{Trainable Parameters of MM-DiT}
We fine-tune both FLUX-dev~\cite{flux1-dev} and SD3.5~\cite{sd3,sd35-8b} by training LoRA~\cite{lora} adapters on specific layers to capture new styles. Multi-Modal Diffusion Transformer (MM-DiT)~\cite{sd3}, the core architecture for both FLUX-dev and SD3.5 comprises dual-stream transformer blocks with separate parameters for text tokens and image tokens, which interact through joint attention mechanisms. To effectively learn the stylistic features encompassing both visual and linguistic characteristics, we train LoRA on the attention layers of \textit{both} modalities. Additionally, FLUX includes single-stream blocks~\cite{vit22b} that handle both modalities simultaneously with attention mechanisms and projection layers that skip this attention, to which we also apply LoRA. This targeted fine-tuning achieves high style-alignment without training the entire network, providing a parameter-efficient method for fine-tuning MM-DiT.

\begin{figure*}[t]
  \centering
   \includegraphics[width=1.0\linewidth]{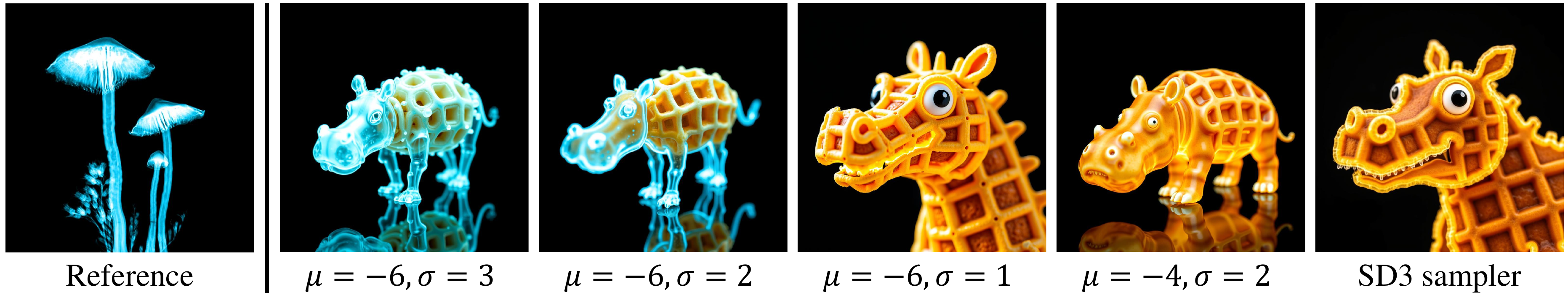}
   \caption{\textbf{Effect of varying $\bm{\mu}$ and $\bm{\sigma}$.} Diffusion models start to capture the reference glowing style when $\mu$ is lower and $\sigma$ is larger. The prompt is `a hybrid creature that is a mix of a waffle and a hippopotamus, in glowing style'. Samples are generated with the same seed.}
   \label{fig:mu_sample}
\end{figure*}

\begin{figure*}
  \centering
  \begin{subfigure}{0.33\linewidth}
    \centering
    \includegraphics[width=1.0\linewidth]{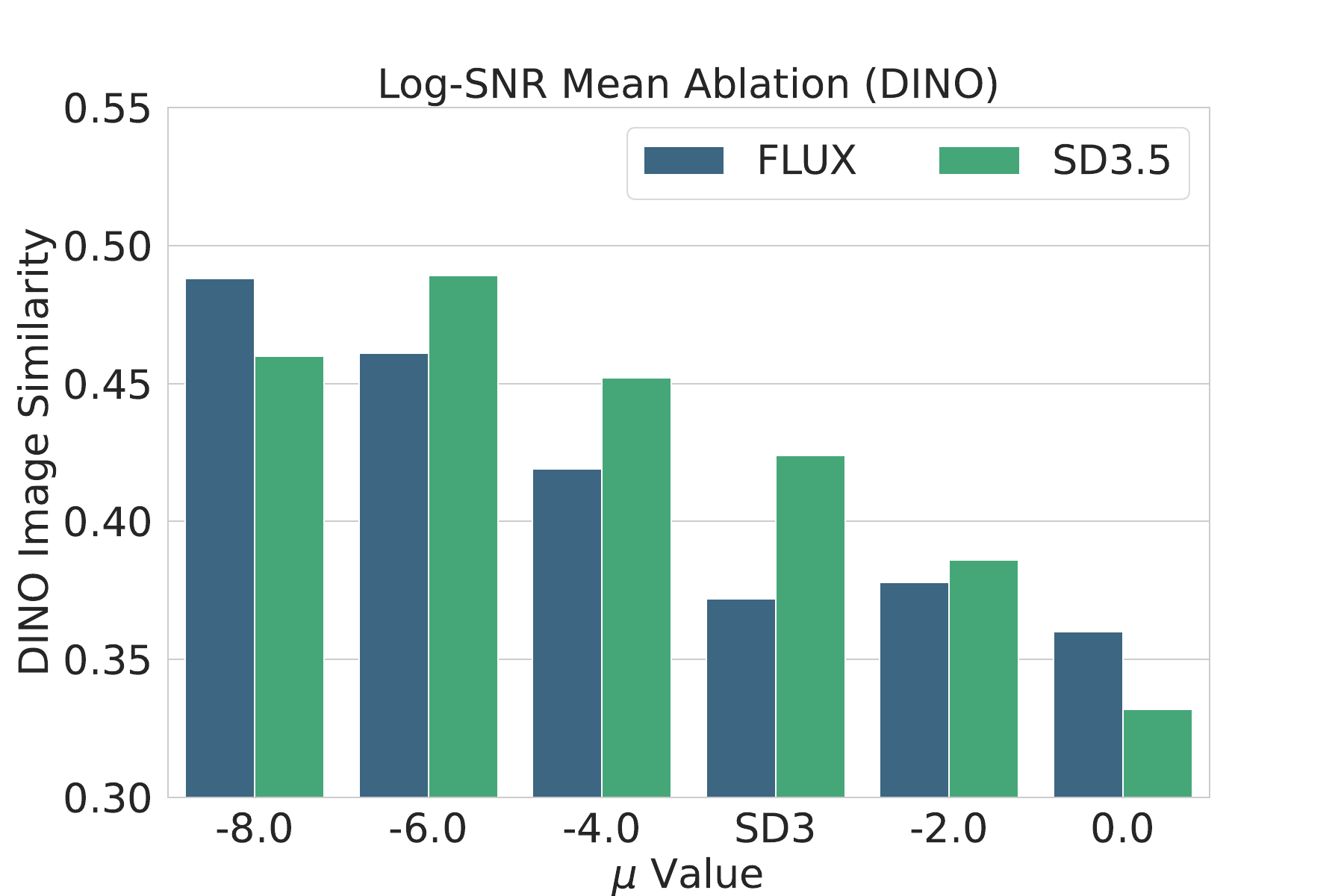}
    \caption{Varying $\mu$.}
    \label{fig:mu_search_dino}
  \end{subfigure}
  \hspace{-1em}
  \begin{subfigure}{0.33\linewidth}
    \centering
    \includegraphics[width=\linewidth]{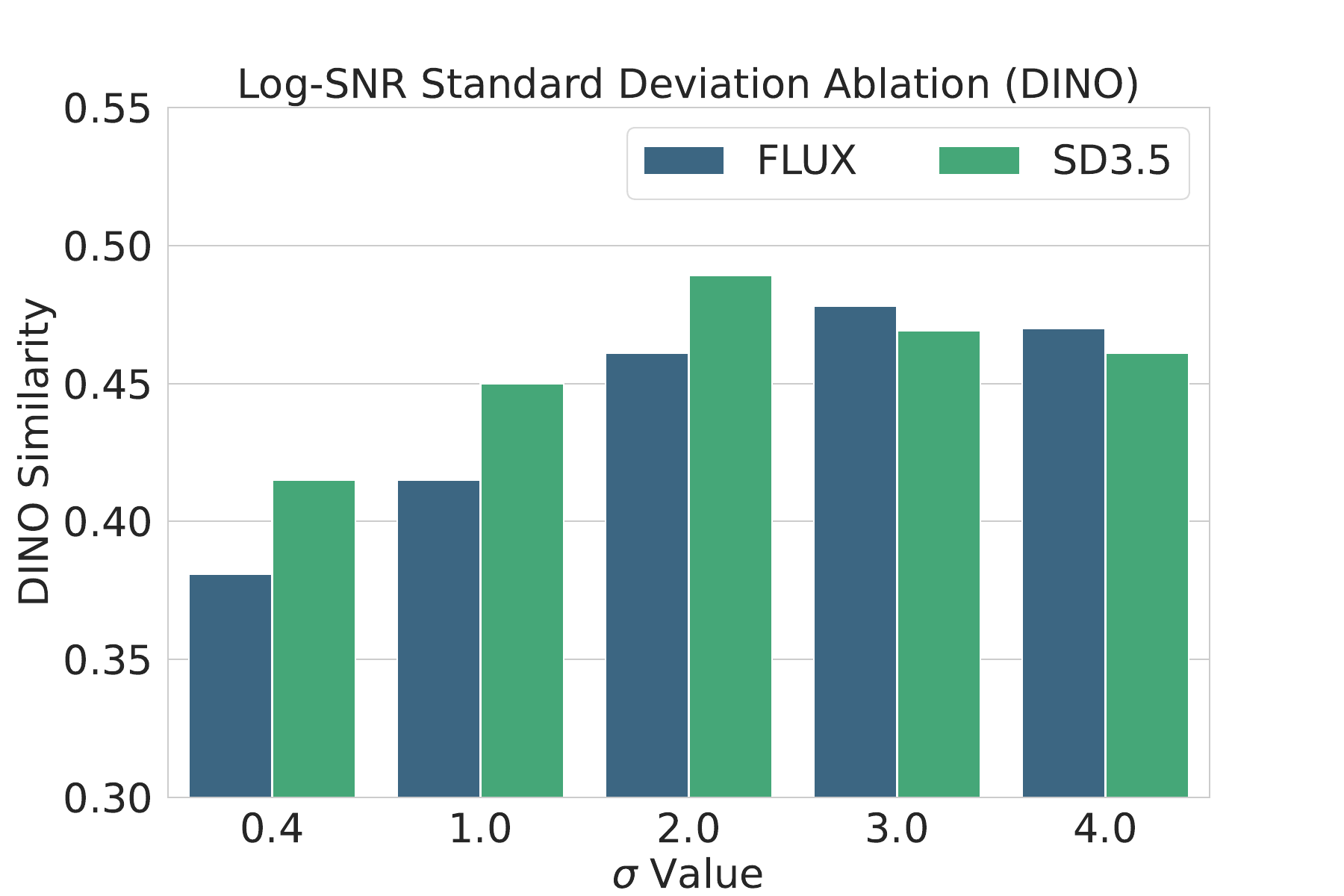}
    \caption{Varying $\sigma$.}
    \label{fig:std_search_dino}
  \end{subfigure}
  \hspace{-1em}
  \begin{subfigure}{0.33\linewidth}
    \centering
    \includegraphics[width=\linewidth]{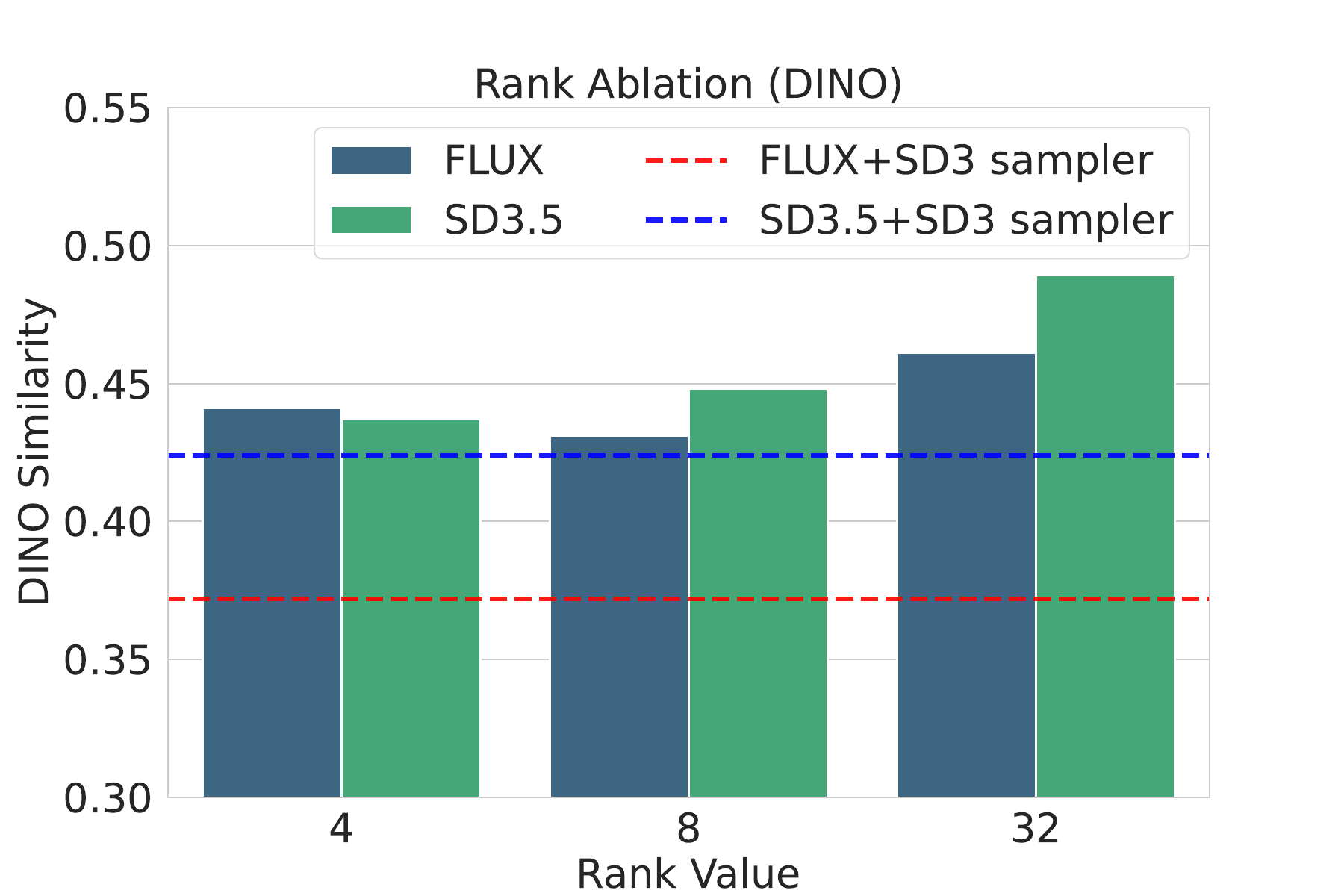}
    \caption{Varying LoRA Rank.}
    \label{fig:rank_search_dino}
  \end{subfigure}
  \caption{\textbf{SNR sampler analysis.} DINO similarities of varying SNR sampler parameters with FLUX and SD3.5-8B. Dotted lines in (c) indicate results of SD3 sampler~\cite{sd3}. Unless specified, we use $\mu=-6$, $\sigma=2$, and rank 32. CLIP scores are shown in the Appendix.}
  \label{fig:search_dino}
\end{figure*}

\section{Experiments}
\label{sec:experiments}
\begin{comment}
We evaluate our method by fine-tuning FLUX-dev~\cite{flux1-dev} and SD3.5-8B~\cite{sd3,sd35-8b} on 18 reference styles from the StyleDrop~\cite{styledrop}. For each reference style, we generate two images for each of the 23 evaluation prompts collected from \cite{styledrop}, resulting in a total of 828 images evaluated per experiment.

For quantitative evaluation, we assess style alignment using DINO~\cite{dino} ViT-S/16 and CLIP~\cite{clip} ViT-B/32 image similarity (CLIP-I), and we measure alignment to the target prompts using CLIP text-image similarity (CLIP-T). 

All models are fine-tuned using the Adam optimizer~\cite{adam} for 300 steps at a learning rate of $10^{-4}$. 
We perform LoRA~\cite{lora} fine-tuning on the frozen pre-trained models, using a rank of 32 in all experiments except for the rank ablation studies. During inference, we use 28 denoising steps and set the guidance~\cite{cfg,meng2023distillation} scale to 7.0.

\end{comment}

% \afterpage{\flushbottom}
\begin{figure*}[t]
  \centering
   \includegraphics[width=\linewidth]{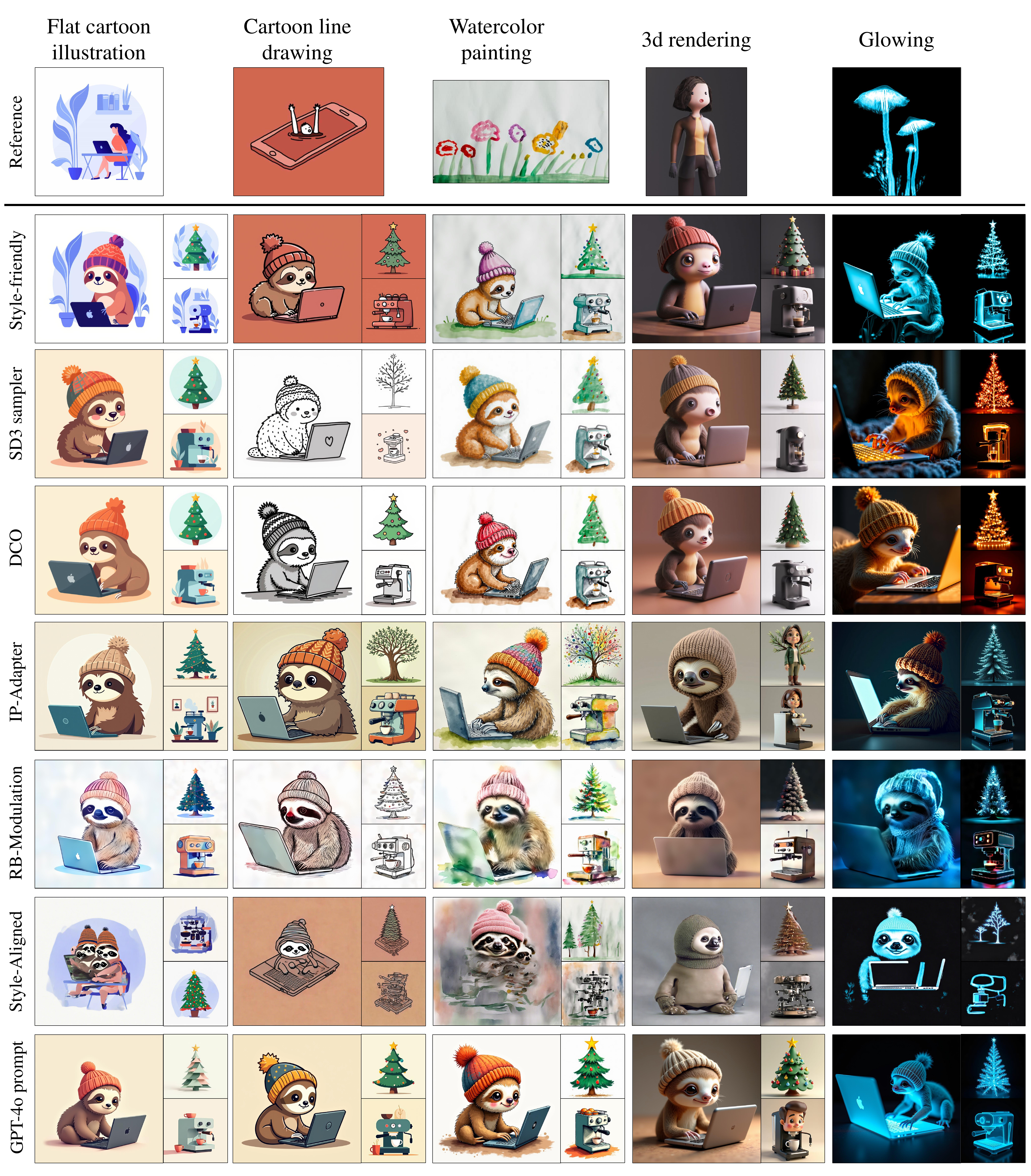}
   \caption{\textbf{Qualitative comparison.} We show `A fluffy baby sloth with a knitted hat trying to figure out a laptop', `A christmas tree', and `An espresso machine' in various styles. Target style prompts are shown above. Our Style-friendly SNR sampler effectively captures the styles indicated by both reference images and their accompanying text prompts. Fine-tuning methods (SD3 sampler and DCO) often miss stylistic nuances due to insufficient focus on the relevant noise levels or overly strong regularization.}
   \label{fig:qualitative_comparison}
\end{figure*}

We compare our method against both fine-tuning and non-fine-tuning approaches for style-driven generation.
\begin{itemize}
    \item \textbf{Fine-tuning}: 
    SD3 sampler \cite{sd3}, Direct Consistency Optimization (DCO) \cite{dco}, and StyleDrop~\cite{styledrop} are selected as representative fine-tuning methods. The SD3 sampler utilizes a flow matching loss~\cite{flowmatching,rectifiedflow} with timestep shifting in \cref{eq:lambda_shift}, while DCO is based on reinforcement learning~\cite{dpo} and employs a regularized loss. For StyleDrop, due to the lack of official open-source implementations, we adopt the unofficial implementation provided by~\cite{styledrop-unofficial}.
    \item \textbf{Image variation}:
    IP-Adapter \cite{ipadapter} and RB-Modulation \cite{rb-modulation} generate image variations using CLIP~\cite{clip} or CSD~\cite{csd} image embeddings respectively. IP-Adapter reconstructs images to best reflect the information contained in the CLIP image embedding, whereas RB-Modulation applies gradient guidance at test time to maximize the similarity score evaluated by CSD.
    \item \textbf{Editing}:
    Style-Aligned \cite{stylealign} is an editing-based method that shares and manually uplifts self-attention weights to enforce the reference style consistency.
    \item \textbf{Detailed prompt}: We name this baseline \textit{GPT-4o Prompt}, where we utilize GPT-4o~\cite{gpt4o} to generate detailed style descriptions from reference images and apply these in text-to-image generation without fine-tuning.
\end{itemize}
Following StyleDrop \cite{styledrop}, we select 18 reference styles as our fine-tuning targets. For each style, we use 23 evaluation prompts, generating 2 images per prompt, which results in a total of 828 images per experiment. We fine-tune FLUX-dev \cite{flux1-dev} and SD3.5-8B \cite{sd3,sd35-8b} with LoRA \cite{lora} at rank 32 (except where noted for rank ablation). Following~\cite{styledrop,stylealign} we measure style alignment via human evaluation, DINO ViT-S/16 \cite{dino}, and CLIP ViT-B/32 \cite{clip} image similarity (CLIP-I), and alignment to the text prompts via CLIP text-image similarity (CLIP-T).
Additional implementation details are in the Appendix.

\subsection{Analysis of Style-Friendly SNR Sampler}
In \cref{fig:mu_sample} and \cref{fig:search_dino}, we conduct experiments to analyze the impact of varying the parameters of our Style-friendly SNR sampler—specifically, the mean ($\mu$) and the standard deviation ($\sigma$) of the log-SNR sampling distribution, as well as the LoRA rank. %We refer to the noise level distribution used for the SD3 pre-training as the SD3 sampler~\cite{sd3}.

\paragraph{Effect of Varying $\bm{\mu}$.}
We experiment with $\mu$ values ranging from 0 to $-8$ for both FLUX-dev and SD3.5-8B. As shown in \cref{fig:mu_sample}, increasing $\mu$ progressively impairs the model's ability to capture reference styles. This trend is quantitatively confirmed in \cref{fig:mu_search_dino}, where DINO similarity scores decrease with increasing $\mu$ (towards zero), clearly indicating poorer style alignment.
Conversely, in \cref{fig:mu_sample}, when $\mu$ is set to $-6$, the models begin to capture and reflect the reference styles effectively.

\paragraph{Effect of Varying $\bm{\sigma}$.}
We also investigate the effect of varying the standard deviation $\sigma$ of the log-SNR sampling distribution in \cref{fig:mu_sample} and \cref{fig:std_search_dino}. When $\sigma$ is less than 2, 
the model does not sufficiently cover the style-emerging region as shown by green curve of \cref{fig:prob_logsnr}, leading to lower style alignment due to limited exposure to critical noise levels. In \cref{fig:mu_sample}, $\sigma\geq 2$ reflects the `glowing' style. Also, in \cref{fig:std_search_dino}, style similarity remains low for $\sigma<2$ but improves when $\sigma\geq 2$. For FLUX, DINO score is highest at $\sigma=3$, whereas for SD3.5, $\sigma=2$ yields the best results. Based on these findings, we adopt these hyperparameters for each respective model.

\paragraph{Effect of Varying Rank.}
In \cref{fig:rank_search_dino}, we examine the impact of model capacity by varying the LoRA rank. Notably, with low $\mu = -6$, a rank of 4 achieves higher DINO similarity compared to the SD3 sampler at rank 32 (dotted lines). This demonstrates focusing on higher noise levels (lower $\lambda_t$) has a more pronounced effect on style learning than model capacity alone.

\subsection{Qualitative Results}
We compare our Style-friendly SNR sampler against previous methods, including the SD3 sampler~\cite{sd3}, Direct Consistency Optimization (DCO)~\cite{dco}, and IP-Adapter~\cite{ipadapter,flux-ip-adapter}, which use FLUX-dev as the backbone model; RB-Modulation~\cite{rb-modulation}, which uses Stable Cascade~\cite{stablecascade}; and Style-Aligned~\cite{stylealign}, which uses SDXL~\cite{sdxl}.

In \cref{fig:qualitative_comparison}, our Style-friendly SNR sampler accurately captures the styles of reference images, reflecting stylistic features including color schemes, layouts, illumination, and brushstrokes.
In contrast, fine-tuning FLUX-dev with the standard SD3 sampler often fails to capture key stylistic components, such as layout (column 1), color scheme (columns 2-4), and illumination (column 5).

Fine-tuning FLUX-dev with DCO struggles to learn the reference styles due to strong regularization that prevents significant deviation from the pre-trained model. As seen in column 4 of the IP-Adapter results, woman in the reference appears, indicating content leakage. IP-Adapter with FLUX-dev and RB-Modulation rely on embeddings of CLIP~\cite{clip} and CSD~\cite{csd}, which may not capture fine stylistic details, leading to less accurate style reproduction. Style-Aligned shares self-attention features within the diffusion model, which can cause artifacts such as destroyed structure (columns 1-3) when attention features conflict.
While GPT-4o prompt utilizes the detailed style description, it often fails to reflect key stylistic features such as color scheme, highlighting the necessity of image guidance for effective style-driven generation.

\subsection{Quantitative Results}
\begin{table}[t]
\setlength{\tabcolsep}{4pt}
  \centering
  \begin{tabular}{l|l|ccc}
  \toprule
     \multicolumn{5}{c}{Style Alignment} \\ \cmidrule{1-5}
   Method    &    Model  & \textit{win}  & \textit{tie}  & \textit{lose}  \\
  \midrule
  Style-Aligned~\cite{stylealign} & SDXL    & \cellcolor{blue!10}61.0 \%       & 7.1\%         & 31.9\%         \\
  RB-Mod~\cite{rb-modulation}              & Cascade         & \cellcolor{blue!10}55.6 \%       & 12.6\%        & 31.8\%         \\
  IP-Adapter~\cite{ipadapter}          & FLUX-dev        & \cellcolor{blue!10}59.2 \%       & 8.0\%         & 32.8\%         \\
  DCO~\cite{dco}                 & FLUX-dev        & \cellcolor{blue!10}56.0 \%       & 10.2\%        & 33.8\%         \\
  SD3 sampler~\cite{sd3}         & FLUX-dev        & \cellcolor{blue!10}56.0 \%       & 9.2\%         & 34.8\%         \\
  \midrule
                        \multicolumn{5}{c}{Text Alignment} \\ \cmidrule{1-5}
    Method     & Model    & \textit{win}  & \textit{tie}  & \textit{lose}  \\
  \midrule
  Style-Aligned~\cite{stylealign}       & SDXL           & \cellcolor{blue!10}60.7\%         & 7.5\%         & 31.8\%         \\
  RB-Mod~\cite{rb-modulation}              & Cascade        & \cellcolor{blue!10}54.3\%         & 6.3\%         & 39.4\%         \\
  IP-Adapter~\cite{ipadapter}          & FLUX-dev       & \cellcolor{blue!10}56.0\%         & 4.6\%         & 39.4\%         \\
  DCO~\cite{dco}                 & FLUX-dev       & \cellcolor{blue!10}53.2\%         & 10.0\%        & 36.8\%         \\
  SD3 sampler~\cite{sd3}         & FLUX-dev       & \cellcolor{blue!10}56.5\%         & 14.0\%        & 29.5\%         \\
  \bottomrule
  \end{tabular}
  \caption{\textbf{Human evaluation.} User preference results comparing style and text alignments between our method and the baselines.}
  \label{tab:user_study}
\end{table}

We further conduct a user study to quantify human preferences using Amazon Mechanical Turk.
Following previous work \cite{styledrop}, we compare our method to each method with two separate questionnaires.
According to the reference style image and target text prompt, users are asked to select which of the two generated images is more similar to the style in the reference image and represents the target text prompt better.
We obtain 450 answers from 150 participants for each comparison, and the results are presented in \cref{tab:user_study}.
Our method outperforms prior arts in both aspects ($p < 0.05$ in the Wilcoxon signed-rank test), consistent with the qualitative results and demonstrates the superiority of our method in learning stylistic elements.
More details on our user study are provided in the Appendix.

In \cref{tab:quantitative_comparison}, we evaluate our method and prior works using DINO~\cite{dino} and CLIP image similarities (CLIP-I) to assess style alignment, and CLIP text-image similarity (CLIP-T) for text alignment. Our method achieves the highest DINO and CLIP-I scores, demonstrating its superior ability to capture styles from the reference images. Notably, our approach improves DCO, indicating that the key factor in style-driven fine-tuning is not whether the method relies on diffusion loss or reinforcement learning-based loss, but rather \textit{which noise levels are emphasized} during training.

While our CLIP-T is slightly lower compared to some methods, we already showed superior text alignment in human evaluation (\cref{tab:user_study}).
This discrepancy arises because textual style descriptions alone can be inherently ambiguous and often correspond to common interpretations. Consequently, methods that accurately reflect unique styles may deviate from these typical interpretations, leading to lower CLIP-T scores despite higher alignment to the intended reference styles.
Overall, our quantitative results confirm that our method accurately reflects both styles and texts.

\begin{table}[t]
\setlength{\tabcolsep}{4pt}
    \centering
    \resizebox{0.48\textwidth}{!}{%
        \begin{tabular}{l|l|ccc}
            \hline
            \multirow{2}{*}{Method} & \multirow{2}{*}{Model} & \multicolumn{3}{c}{Metrics} \\
            \cline{3-5}
            & & DINO $\uparrow$ & CLIP-I $\uparrow$ & CLIP-T $\uparrow$ \\
            \hline
            Style-Aligned~\cite{stylealign} & SDXL & 0.410 & 0.675 & 0.340 \\
            RB-Mod~\cite{rb-modulation} & Cascade & 0.317 & 0.647 & 0.363 \\
            IP-Adapter~\cite{ipadapter} & FLUX & 0.361 & 0.656 & 0.354 \\
            GPT-4o Prompt & FLUX & 0.299 & 0.621 & 0.338 \\
            \hline
            StyleDrop~\cite{styledrop}\textsuperscript{\textdagger} & MUSE & 0.465 & 0.665 & 0.325 \\
            \hline
            SD3 sampler~\cite{sd3} & SD3.5 & 0.424 & 0.670 & 0.350 \\
            ~~~+\textbf{Style-friendly} & SD3.5 & \cellcolor{blue!10}0.489 & \cellcolor{blue!10}0.698 & 0.349 \\
            DCO~\cite{dco} & SD3.5 & 0.399 & 0.661 & \cellcolor{blue!10}0.355 \\
            ~~~+\textbf{Style-friendly} & SD3.5 & 0.478 & 0.695 & 0.351 \\
            \hline
            SD3 sampler~\cite{sd3} & FLUX & 0.373 & 0.645 & 0.350 \\
            ~~~+\textbf{Style-friendly} & FLUX & 0.478 & 0.691 & 0.343 \\
            DCO~\cite{dco} & FLUX & 0.373 & 0.643 & \cellcolor{blue!10}0.353 \\
            ~~~+\textbf{Style-friendly} & FLUX & \cellcolor{blue!10}0.488 & \cellcolor{blue!10}0.698 & 0.341 \\
            \hline
        \end{tabular}%
    }
    \caption{\textbf{Quantitative comparison.} Style alignment (DINO and CLIP-I) and text alignment (CLIP-T) with 18 styles from \cite{styledrop}. Our style-friendly exhibits superior style-alignment scores. Rows 1-3 show non-fine-tuning baselines. \textdagger: Results obtained using an unofficial implementation~\cite{styledrop-unofficial}.}
    \vspace{-1em}
    \label{tab:quantitative_comparison}
\end{table}

\subsection{Applications}

\begin{figure}[t]
  \centering
   \includegraphics[width=\linewidth]{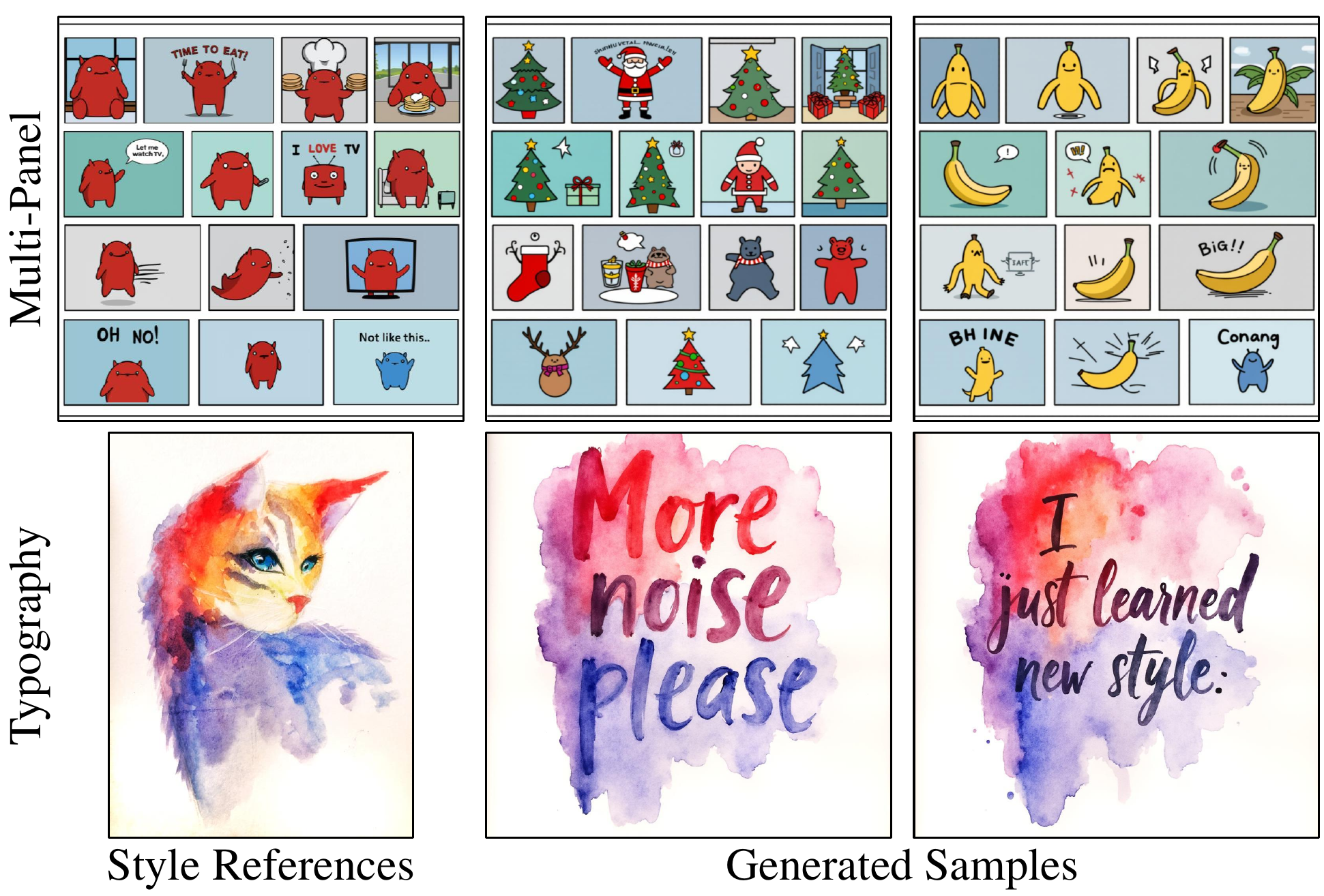}
   \caption{\textbf{Multi-panel and typography.} First row demonstrates generating multiple coherent panels as a \textit{single image}. Second row shows customized typography with a unique style.}
   \label{fig:application}
\end{figure}

While Dreambooth~\cite{dreambooth} paper demonstrates generating multi-panel comics by generating \textit{each panel} with a fine-tuned diffusion model, we define the entire multi-panel comic itself as a unique style.  Our method treats multiple panels as a \textit{single image} during fine-tuning, enabling the generation of coherent multi-panel comics from only a single reference (see the first row of \cref{fig:application}). By specifying a new subject in the target prompt, the model consistently places that subject across all comic-style panels. Beyond comics, our method also extends to typography, leveraging the spelling capabilities of recent models~\cite{flux1-dev,sd35-8b}. 
As shown in the second row of \cref{fig:application}, this flexibility allows users to effortlessly generate a broad range of customized textual elements in unique styles.

\section{Related Works}
\label{sec:related}

\subsection{Diffusion Models}

Diffusion models generate data from noise, encapsulating approaches based on denoising score matching~\cite{ncsn,sde,edm}, maximum likelihood training~\cite{vdm}, and rectified flow~\cite{flowmatching,rectifiedflow}. 
One of the critical factors influencing the performance of diffusion models is the sampling distribution over noise levels during training, known as the \textit{importance sampling} of noise levels. Studies focusing on noise schedule adjustment~\cite{simple,iddpm} and weight adjustment~\cite{p2,edm,understanding} have succeeded in training high-quality diffusion models by carefully weighting different noise levels. Their effectiveness has been shown with object-centric metrics and benchmarks~\cite{fid,kynkaanniemirole,t2icompbench,geneval}.

\subsection{Style-Driven Generation}
With advancements in text-to-image models, practitioners have increasingly sought to generate images featuring personal styles~\cite{styledrop,ipadapter,rb-modulation,dco}. Fine-tuning methods~\cite{dreambooth,styledrop,dco} have been particularly prominent in this area. StyleDrop~\cite{styledrop}, a study closely related to our work, utilizes a masked generative model~\cite{muse} and involves human data selection through multi-stage training. Some works focus on learning multiple concepts simultaneously~\cite{custom,instruct-imagen} or merging several fine-tuned models~\cite{ziplora,dco}, while others analyze the diffusion model's U-Net~\cite{unet} layers to identify those most effective for learning styles ~\cite{b-lora}.
%However, fine-tuning requires hyperparameter searches with each new large-scale model release~\cite{sdxl,sd3,flux1-dev}, and is often applied without a deep understanding of the diffusion objectives. 
%However, several methods are not readily applicable to newly released large-scale models~\cite{sd3,flux1-dev}, limiting their scalability as text-to-image models evolve. In contrast, our method provides a scalable and adaptable fine-tuning strategy, validated on up-to-date diffusion models.
However, several existing methods have not yet been validated or applied to recently released large-scale models~\cite{sd3,flux1-dev}. In contrast, our method provides an adaptable fine-tuning strategy, which we validate on up-to-date diffusion models.

As an alternative approach for style-driven generation, zero-shot approaches have been proposed~\cite{ipadapter,rb-modulation,stylealign,instantstyle}, but these methods still fall short in style alignment compared to fine-tuning and are often limited to specific domains~\cite{imagineyourself,hyperdreambooth}. Furthermore, some methods~\cite{rb-modulation,stylealign} require extra inference-time gradient guidance~\cite{rb-modulation} or inversion~\cite{stylealign}, increasing inference cost. Due to these limitations, we focus on fine-tuning in our work, aiming to provide insights into the behavior of diffusion objectives to make fine-tuning more accessible and effective. While fine-tuning is not the only option, our results show that it is a promising approach.
\section{Conclusion}
\label{sec:conclusion}

In this paper, we observed that stylistic features in diffusion models emerge predominantly at higher noise levels. To address the limitations of previous fine-tuning approaches in capturing new artistic styles, we proposed the \textit{Style-friendly SNR sampler}, which biases the SNR distribution towards higher noise levels. We showed style-driven generation that reflects reference styles specified via text prompts. We hope this work will serve as a stepping stone toward using diffusion models as digital art previewers. 

\paragraph{Acknowledgement} This work was supported by the National Research Foundation of Korea (NRF) grant funded by the Korea government (MSIT) [No. 2022R1A3B1077720], Institute of Information \& communications Technology Planning \& Evaluation (IITP) grant funded by the Korea government(MSIT) [NO.RS-2021-II211343, Artificial Intelligence Graduate School Program (Seoul National University)], and the BK21 FOUR program of the Education and Research Program for Future ICT Pioneers, Seoul National University in 2024. 

{
    \small
    \bibliographystyle{ieeenat_fullname}
    \bibliography{main}
}
\clearpage
\appendix
\setcounter{section}{0}
\renewcommand\thesection{\Alph{section}}
\setcounter{table}{0}
\renewcommand{\thetable}{S\arabic{table}}
\setcounter{figure}{0}
\renewcommand{\thefigure}{S\arabic{figure}}
\maketitlesupplementary
%\section*{\Large{Appendix}}

\section{Experimental Details}
\label{sec:exp_detail}

\subsection{Style Prompts}
\label{sec:style_prompts}
We conduct all quantitative evaluations using the 18 reference styles shown in the appendix of the StyleDrop paper~\cite{styledrop}. The style prompts for these 18 styles can also be found in the StyleDrop appendix. 
%For qualitative evaluations, we use additional challenging style references and we display the corresponding style prompt for each image in \cref{fig:data}. The first and second references are produced with open-source meme generator \href{https://imgflip.com/memegenerator/31945629/You-Just-Activated-My-Trap-Card}{source1} and \href{https://imgflip.com/memegenerator/Drake-Hotline-Bling}{source2} respectively. The third reference is cropped from the appendix of Dreambooth~\cite{dreambooth} paper.

\begin{comment}
\begin{figure}
  \centering
   \includegraphics[width=1.0\linewidth]{fig/data.pdf}
   \caption{\textbf{Style prompts used for experiments.}}
   \label{fig:data}
\end{figure}
\end{comment}

\subsection{Evaluation Prompts}
We present 23 evaluation prompts collected from StyleDrop paper~\cite{styledrop} used for our quantitative and qualitative comparisons: 
\begin{itemize}
    \item An Opera house in Sydney in \{style prompt\} style
    \item A fluffy baby sloth with a knitted hat trying to figure out a laptop, close up in \{style prompt\} style
    \item A Golden Gate bridge in \{style prompt\} style
    \item The letter `G' in \{style prompt\} style
    \item A man riding a snowboard in \{style prompt\} style
    \item A panda eating bamboo in \{style prompt\} style
    \item A friendly robot in \{style prompt\} style
    \item A baby penguin in \{style prompt\} style
    \item A moose in \{style prompt\} style
    \item A towel in \{style prompt\} style
    \item An espresso machine in \{style prompt\} style
    \item An avocado in \{style prompt\} style
    \item A crown in \{style prompt\} style
    \item A banana in \{style prompt\} style
    \item A bench in \{style prompt\} style
    \item A boat in \{style prompt\} style
    \item A butterfly in \{style prompt\} style
    \item An F1 race car in \{style prompt\} style
    \item A Christmas tree in \{style prompt\} style
    \item A cow in \{style prompt\} style
    \item A hat in \{style prompt\} style
    \item A piano in \{style prompt\} style
    \item A wood cabin in \{style prompt\} style
\end{itemize}

\begin{figure*}[t]
  \centering
   \includegraphics[width=1.0\linewidth]{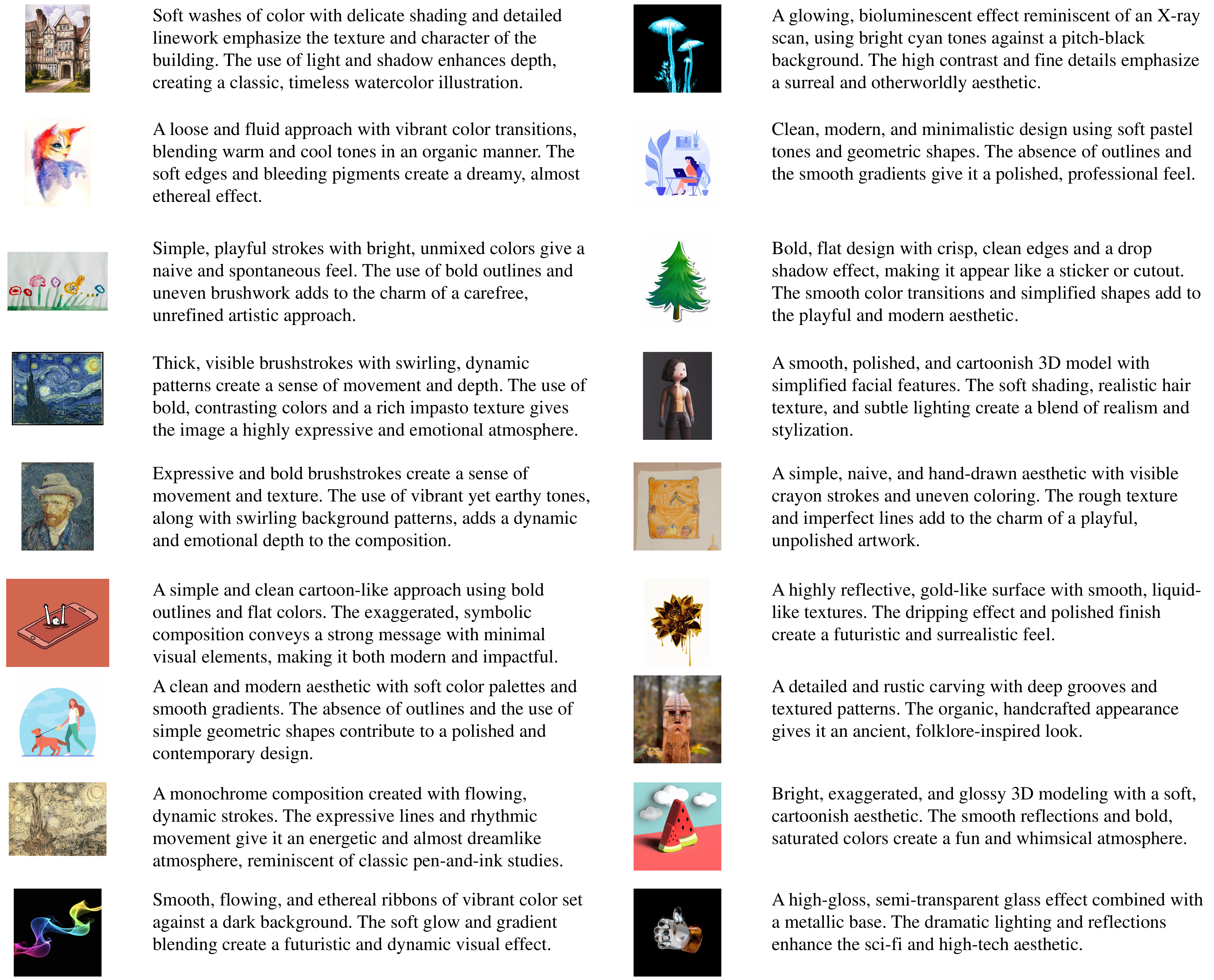}
   \caption{\textbf{Detailed style description.} These descriptions serve as the input prompts for our GPT-4o~\cite{gpt4o} prompt baseline, which generates images solely based on these textual style specifications.}
   \label{fig:gpt4o}
\end{figure*}

In \cref{fig:gpt4o}, we present the detailed style descriptions generated by GPT-4o, which serve as text prompts for the \textit{GPT-4o Prompt} baseline. Specifically, for each of the 18 reference styles, we obtained comprehensive textual descriptions using GPT-4o. These detailed prompts were directly used for text-to-image generation.

\subsection{User Study}
\label{sec:user}
In this section, we provide detailed information about the setup of our user study.
Our user study aims to measure human preferences in two key objectives of style-driven image generation: style alignment and text alignment.
To assess these preferences, we conduct pairwise comparisons between our method and each baseline for each objective.
Participants are shown the reference image, target text prompt, and two generated images (one from each method) and are asked to choose the image that better satisfies the objective.
We collect three responses from each of the 150 participants, resulting in a total of 450 responses for each comparison. 
The full instructions used in our questionnaires are as follows.

For style alignment objective,
\begin{itemize}[leftmargin=3em]
    \item Given a reference image and two machine-generated images, select which machine-generated output better matches the style of the reference image for each pair.
    \item Please focus only on the style including color schemes, layouts, illumination, and brushstrokes.
    \item If it’s difficult to determine a preference, please select ``Cannot Determine / Both Equally".
\end{itemize}

For text alignment objective,
\begin{itemize}[leftmargin=3em]
    \item Given a reference image and two machine-generated images, select which machine-generated output better matches the target text for each pair.
    \item Please focus only on the text, without regard for the reference image.
    \item If it’s difficult to determine a preference, please select ``Cannot Determine / Both Equally".
\end{itemize}

\subsection{Implementation}
To ensure reproducibility, we provide pseudo-code implementations of Style-friendly SNR samplers in \cref{fig:code_snr} and the addition of LoRA~\cite{lora} parameters to MM-DiT for training in \cref{fig:code_lora}.

\textbf{Optimizer and Learning Rate.}
We use the Adam optimizer \cite{adam} at a learning rate of 10\textsuperscript{-4} for 300 steps. The batch size is set to 1, with gradient accumulation over 4 steps.

\textbf{Guidance Scale and Inference Steps.}
During inference, we use a guidance scale \cite{cfg,meng2023distillation} of 7.0. The number of denoising steps is set to 28.

\textbf{Model-Specific Details.} FLUX-dev~\cite{flux1-dev} is a guidance-distilled model \cite{meng2023distillation} that takes guidance scale as input. For fine-tuning, we fix the guidance scale to 1.0 to match standard diffusion training, enable gradient checkpointing for memory efficiency, and use BF16 quantization for both fine-tuning and inference. SD3.5-8B~\cite{sd35-8b} use FP16 precision.

\textbf{Disabling Timestep Shifting.}
SD3~\cite{sd3} uses a timestep shifting mechanism. However, we disable this shifting for our Style-friendly SNR sampler to isolate the effect of our proposed SNR sampling strategy.

\textbf{Baseline Implementation.}
We use the Hugging Face Diffusers library (version 0.31.0) for consistent training and inference across methods. 

\begin{figure*}[t]
\centering
\begin{minipage}{\textwidth} % Use minipage to control width
\begin{pythoncode}
# Inputs: mu, sigma, B, latent
# sample log-SNR
logsnr = torch.normal(mean=mu, std=sigma, size=(B,))
# compute timestep t
t = torch.nn.functional.sigmoid(-logsnr / 2).view(B, 1, 1, 1)  
# sample noise
noise = torch.randn_like(latent)
# diffuse latent
noisy_latent = (1.0 - t) * latent + t * noise
\end{pythoncode}
\end{minipage}
\caption{\textbf{PyTorch implementation of a Style-friendly SNR sampler.}}
\label{fig:code_snr}
\end{figure*}

\begin{figure*}[t]
\centering
\begin{minipage}{\textwidth}
\begin{pythoncode}
# Inputs: model_name, rank
# Configure LoRA for the specified model
if model_name == "FLUX":
    target_modules = [
        "to_k", "to_q", "to_v", "to_out.0",
        "add_k_proj", "add_q_proj", "add_v_proj", "proj_mlp", "proj_out"
    ]
elif model_name == "SD3":
    target_modules = [
        "to_k", "to_q", "to_v", "to_out.0",
        "add_k_proj", "add_q_proj", "add_v_proj", "to_add_out"
    ]
else:
    raise ValueError(f"Unsupported model: {model_name}")

# LoRA configuration
transformer_lora_config = LoraConfig(
    r=rank,
    lora_alpha=rank,
    init_lora_weights="gaussian",
    target_modules=target_modules,
)

# Add adapter to the transformer
transformer.add_adapter(transformer_lora_config)
\end{pythoncode}
\end{minipage}
\caption{\textbf{PyTorch implementation of LoRA integration.}}
\label{fig:code_lora}
\end{figure*}

\section{Baselines}

\subsection{Direct Consistency Optimization}

Direct Consistency Optimization (DCO)~\cite{dco} is a fine-tuning method inspired by direct preference optimization~\cite{dpo} commonly used in large language models (LLMs). Instead of directly minimizing the diffusion loss, DCO aims to ensure that the diffusion loss of the fine-tuned model is lower than that of the pre-trained model on the reference data. The objective function is defined as:
\begin{align}
    &\mathcal{L}_{DCO}(x_0) = \mathbb{E}_{t,\epsilon}\bigg[-\text{log}\sigma(-\beta T \notag \\ &||v_{\theta}(x_t,t)-v(x_t,t)||^2 -||v_{\phi}(x_t,t)-v(x_t,t)||^2)\bigg],
  \label{eq:dco}
\end{align}
where $v(x_t,t)$ is target velocity field, $v_{\theta}$, is fine-tuning model, and $v_{\phi}$ is frozen pre-trained model.

In this objective, the parameter $\beta T$ controls the strength of the preference towards the fine-tuned model over the pre-trained model. DCO increases the relative likelihood of the fine-tuned model over the pre-trained model, penalizing less when the fine-tuned model's loss is smaller. This helps preserve the text-to-image alignment of the pre-trained model.

However, DCO requires computations involving both the fine-tuned and pre-trained models, making it computationally more intensive than directly fine-tuning using the standard diffusion loss. In our experiments, we observed that using a large value of $\beta T=1000$ resulted in slower convergence and suboptimal performance. Therefore, we set $\beta T = 1$ to achieve better results.

\subsection{IP-Adapter}
IP-Adapter~\cite{ipadapter} is designed to enable text-to-image models to generate identity-preserving images by training a compact adapter that encodes CLIP image embeddings~\cite{clip}. This adapter introduces the CLIP image embedding as an additional input by concatenating its output with the text embeddings. The parameter-efficient nature of IP-Adapter allows for easy training and deployment across various text-to-image models. However, a notable limitation is its restricted style alignment due to the expressive constraints of CLIP embeddings, which may result in generated images that do not fully capture detailed stylistic characteristics. IP-Adapter allows adjusting the conditioning strength by scaling the embeddings with a factor between 0 and 1; we use a scale of 0.6 in all experiments. Using a scale of 1 can lead to content leakage beyond the style.

\subsection{RB-Modulation}

RB-Modulation~\cite{rb-modulation} is a zero-shot approach using Stable Cascade~\cite{stablecascade}, a model accepting both CLIP image embeddings and text embeddings as inputs. During the denoising process, RB-Modulation employs gradient guidance of a CSD~\cite{csd}, a model fine-tuned from CLIP to measure style similarity, resembling classifier guidance~\cite{guided-diffusion}. At each denoising step, CSD computes the similarity between the approximated $x_0$ and the reference image, guiding the generation process to enhance this similarity. RB-Modulation also aggregates multiple attention features.

However, this approach relies on models that accept CLIP image embeddings, limiting model selection. Additionally, using gradient guidance of CSD increases inference costs, making the generation process more computationally intensive.

\subsection{Style-Aligned}
Style-Aligned~\cite{stylealign} generates consistent sets of images with the same style by ensuring that features of each image attend to those of a reference image through shared key and value features in self-attention layers of image tokens. It first maps the reference image to noise using DDIM inversion~\cite{ddim} and shares self-attention features during denoising. The fidelity to the reference style can be controlled by amplifying the self-attention logits in the diffusion model. However, Style-Aligned is not directly applicable to MM-DiT~\cite{sd3} architecture that lacks image-only self-attention layers. Moreover, artificially amplifying self-attention logits can lead to artifacts and lower-quality images due to conflicting attention features.

\subsection{Offset Noise}
Offset noise~\cite{offset} is a method proposed to fine-tune diffusion models for generating monochromatic images. During the diffusion process, a constant offset noise—identical across all pixel positions—is added to the standard Gaussian noise, scaled by a small factor (e.g., 0.1). This introduces an explicit bias toward monotonic noise patterns, encouraging the model to learn and reproduce solid colors. While offset noise aids in learning monotonous patterns, it can hinder the model's capacity to learn more complex styles.

Here, we additionally experiment with incorporating offset noise into our training process in \cref{tab:offset_add}. Offset noise with a scale of 0.1 improves the SD3 sampler's results in DINO and CLIP-I scores, as many reference styles from the StyleDrop paper~\cite{styledrop} have monochromatic backgrounds, favoring this trick. However, it still does not reach the performance of our Style-friendly SNR sampler. Moreover, when we combine our Style-friendly approach with a smaller scale of offset noise (0.01), we observe a slight improvement in the style alignment of FLUX-dev.

This quantitative evaluation is based on the monochromatic backgrounds prevalent in the StyleDrop~\cite{styledrop} references. Our qualitative comparisons in \cref{fig:template_comparison} show that offset noise struggles with complex references, failing to capture intricate stylistic details. This indicates that while offset noise can help with simple, uniform styles, it is vulnerable to complex styles.

\begin{table}[t]
    \setlength{\tabcolsep}{4pt}
    \centering
    \resizebox{0.47\textwidth}{!}{%
        \begin{tabular}{l|l|ccc}
            \hline
            \multirow{2}{*}{Method} & \multirow{2}{*}{Model} & \multicolumn{3}{c}{Metrics} \\
            \cline{3-5}
            & & DINO $\uparrow$ & CLIP-I $\uparrow$ & CLIP-T $\uparrow$ \\
            \hline
            SD3 Sampler~\cite{sd3} & SD3.5 & 0.424 & 0.670 & 0.350 \\
            ~~~~w/ offset 0.1  & SD3.5 & 0.452 & 0.678 & \cellcolor{blue!10}0.353 \\
            \textbf{Style-friendly} & SD3.5 & \cellcolor{blue!10}0.489 & \cellcolor{blue!10}0.698 & 0.349 \\
            ~~~~w/ offset 0.01  & SD3.5 & 0.476 & 0.697 & 0.350 \\
            \hline
            SD3 Sampler~\cite{sd3} & FLUX-dev & 0.373 & 0.645 & \cellcolor{blue!10}0.350 \\
            ~~~~w/ offset 0.1  & FLUX-dev & 0.451 & 0.679 & 0.349 \\
            \textbf{Style-friendly} & FLUX-dev & 0.461 & 0.686 & 0.344 \\
            ~~~~w/ offset 0.01  & FLUX-dev & \cellcolor{blue!10}0.500 & \cellcolor{blue!10}0.704 & 0.341 \\
            \hline
        \end{tabular}%
    }
    \caption{\textbf{Incorporating offset noise.} Offset noise improves SD3 sampler but still does not reach the performance of our Style-friendly SNR sampler; combining our Style-friendly approach with Offset Noise at a smaller scale (0.01) slightly enhances the style alignment of FLUX-dev. Here, we use $\sigma=2$ for Style-friendly.}
    \label{tab:offset_add}
\end{table}

\section{Additional Results}

\subsection{Quantitative Results}
%%%% QUANTITATIVE FIGURES %%%%
\begin{figure*}[t]
  \centering
  \begin{subfigure}{0.33\linewidth}
    \centering
    \includegraphics[width=\linewidth]{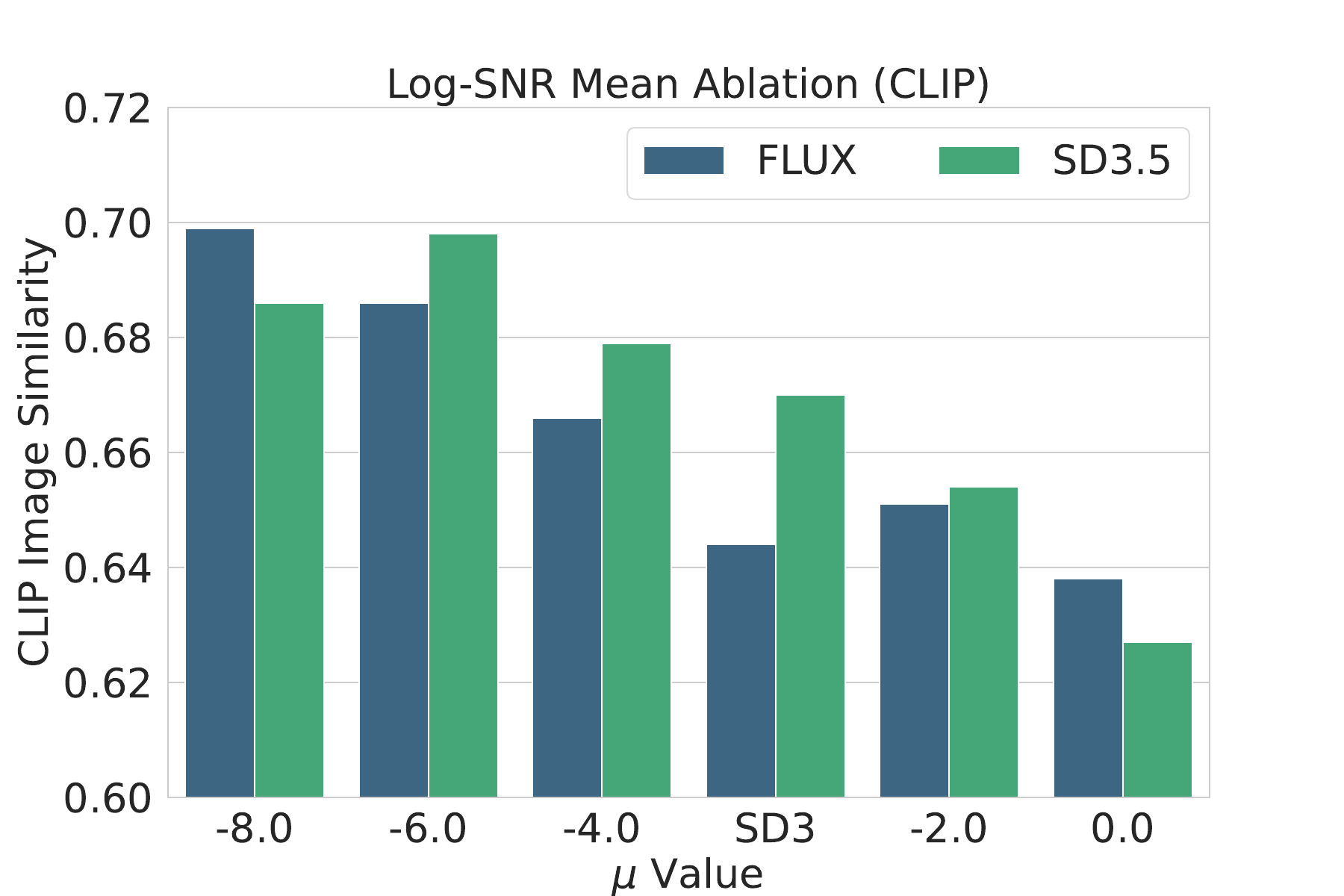}
    \caption{Varying $\mu$.}
    \label{fig:mu_search_clip}
  \end{subfigure}
  \hspace{-1em}
  \begin{subfigure}{0.33\linewidth}
    \centering
    \includegraphics[width=\linewidth]{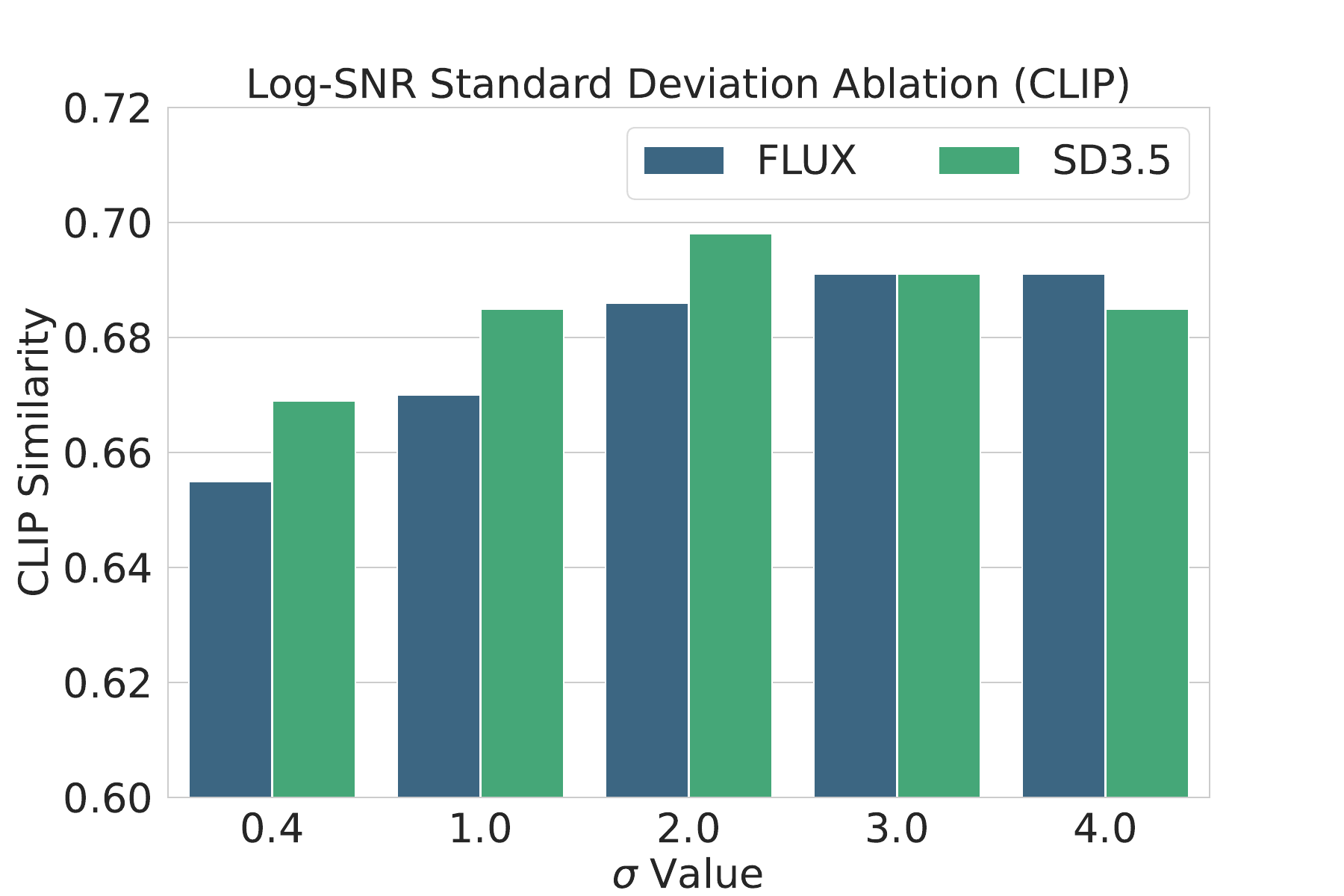}
    \caption{Varying $\sigma$.}
    \label{fig:std_search_clip}
  \end{subfigure}
  \hspace{-1em}
  \begin{subfigure}{0.33\linewidth}
    \centering
    \includegraphics[width=\linewidth]{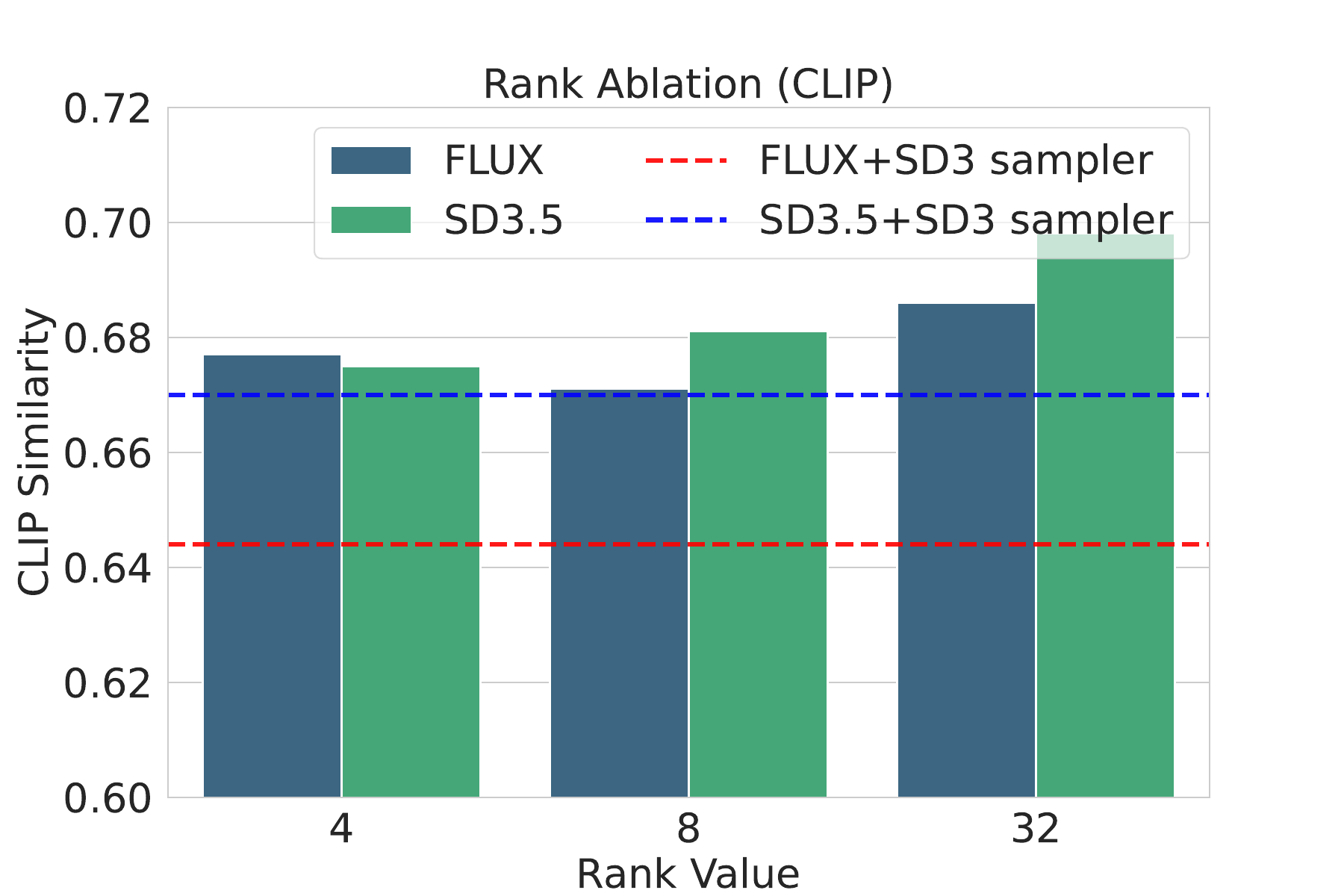}
    \caption{Varying LoRA Rank.}
    \label{fig:rank_search_clip}
  \end{subfigure}
  \caption{\textbf{SNR sampler analysis.} CLIP-I similarities with FLUX and SD3.5-8B. Dotted lines in (c) indicate the results of SD3 sampler~\cite{sd3}.}
  \label{fig:search_clip}
\end{figure*}

\paragraph{CLIP Scores.}
In the main paper, we presented analyses of the mean $\mu$, standard deviation $\sigma$, and LoRA rank using the DINO similarity score. In \cref{fig:mu_search_clip}, we provide the corresponding CLIP image similarity (CLIP-I) scores to further validate our findings. The CLIP-I scores exhibit a similar trend to the DINO scores, where decreasing $\mu$ enhances style alignment. Varying $\sigma$ affects the CLIP-I scores consistently with the DINO results. Our Style-friendly SNR sampler with $\mu = -6$ and a rank of 4 still outperforms the SD3 sampler with a rank of 32 (dotted lines). 

%%%%%%% TABLES %%%%%%%%
\begin{table}[t]
    \setlength{\tabcolsep}{4pt}
    \centering
    \resizebox{0.47\textwidth}{!}{%
        \begin{tabular}{l|l|ccc}
            \hline
            \multirow{2}{*}{Method} & \multirow{2}{*}{Model} & \multicolumn{3}{c}{Metrics} \\
            \cline{3-5}
            & & DINO $\uparrow$ & CLIP-I $\uparrow$ & CLIP-T $\uparrow$ \\
            \hline
            SD3 Sampler~\cite{sd3} & FLUX-dev & 0.373 & 0.645 & \cellcolor{blue!10}0.350 \\
            ~~~~w/ rank 128  & FLUX-dev & 0.426 & 0.668 & 0.345 \\
            \textbf{Style-friendly} & FLUX-dev & \cellcolor{blue!10}0.461 & \cellcolor{blue!10}0.686 & 0.344 \\
            \hline
        \end{tabular}%
    }
    \caption{\textbf{Comparison to increasing LoRA rank.}}
    \label{tab:rank_large}
\end{table}
\begin{table}[t]
    \centering
    \begin{tabular}{l|ccc}
        \toprule
        Method & DINO & CLIP-I & CLIP-T \\
        \midrule
        Style-friendly & \cellcolor{blue!10}0.489 & \cellcolor{blue!10}0.698 & \cellcolor{blue!10}0.349 \\
        ~w/o Text attn & 0.462 & 0.693 & \cellcolor{blue!10}0.349 \\
        \bottomrule
    \end{tabular}
    \caption{\textbf{Ablation study on trainable parameters.}}
    \label{tab:parameter}
\end{table}

\paragraph{Effectiveness Compared to Increasing Model Capacity.}
To demonstrate that our method is more effective than increasing model capacity, we conduct an additional experiment where we fine-tune the model using the SD3 sampler with a higher LoRA rank of 128. As shown in \cref{tab:rank_large}, our Style-friendly SNR sampler with a rank of 32 achieves higher DINO and CLIP-I scores compared to the SD3 sampler with a rank of 128. This indicates that focusing on the critical noise levels where styles emerge has a more significant impact than increasing the number of trainable parameters.

\paragraph{Trainable Parameters.}
To validate the importance of fine-tuning both transformer blocks of MM-DiT~\cite{sd3}, we conduct an ablation study on SD3.5-8B, comparing the results of training LoRA adapters on only the image-transformer blocks versus training on both the image and text-transformer blocks. As shown in \cref{tab:parameter}, fine-tuning both the image and text-transformer blocks leads to higher DINO and CLIP-I scores compared to fine-tuning only the image-transformer blocks, while the CLIP-T scores are identical. This indicates that including the text-transformer blocks in the fine-tuning process enhances the model's ability to learn stylistic features without compromising text alignment. These results suggest that to effectively capture new styles, it is beneficial to fine-tune both the visual and linguistic components of MM-DiT.

\subsection{Qualitative Results}

\begin{figure*}[t]
  \centering
   \includegraphics[width=1.0\linewidth]{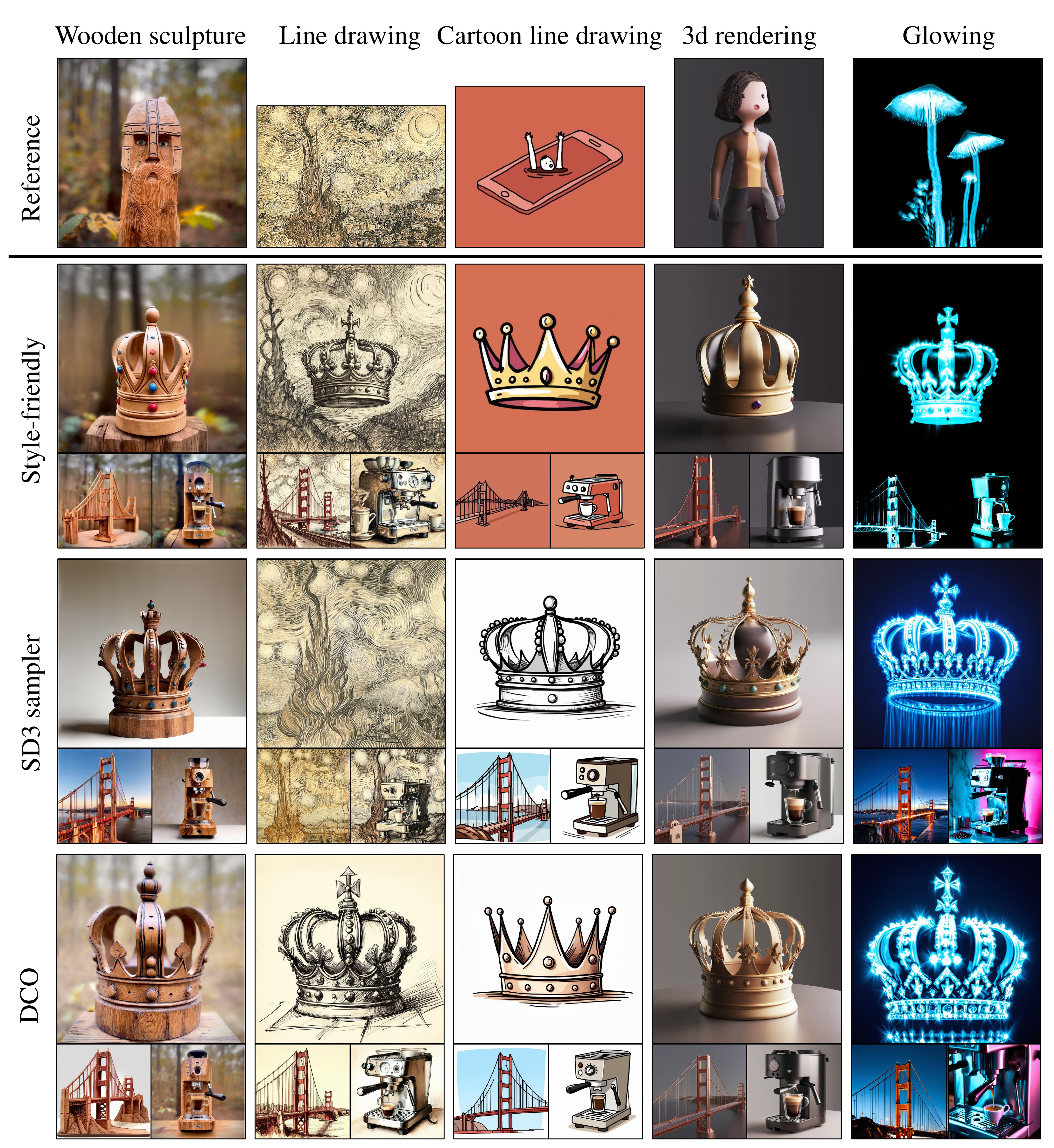}
   \caption{\textbf{Comparison of fine-tuning the SD3.5-8B.} We show `A crown', `A Golden Gate bridge', and `An espresso machine' in various styles. The results with SD3.5-8B are consistent with the qualitative comparison based on FLUX-dev presented in the main paper.}
   \label{fig:compare_sd35}
\end{figure*}

\paragraph{SD3.5 Samples.}
We extend our qualitative comparison by evaluating our Style-friendly SNR sampler using the SD3.5-8B model~\cite{sd35-8b}, comparing it against previous fine-tuning methods, namely the SD3 sampler~\cite{sd3} and DCO~\cite{dco}. As shown in \cref{fig:compare_sd35}, the results are consistent with the qualitative comparisons using FLUX-dev presented in the main paper.

\begin{figure*}
    \centering
    \includegraphics[width=\linewidth]{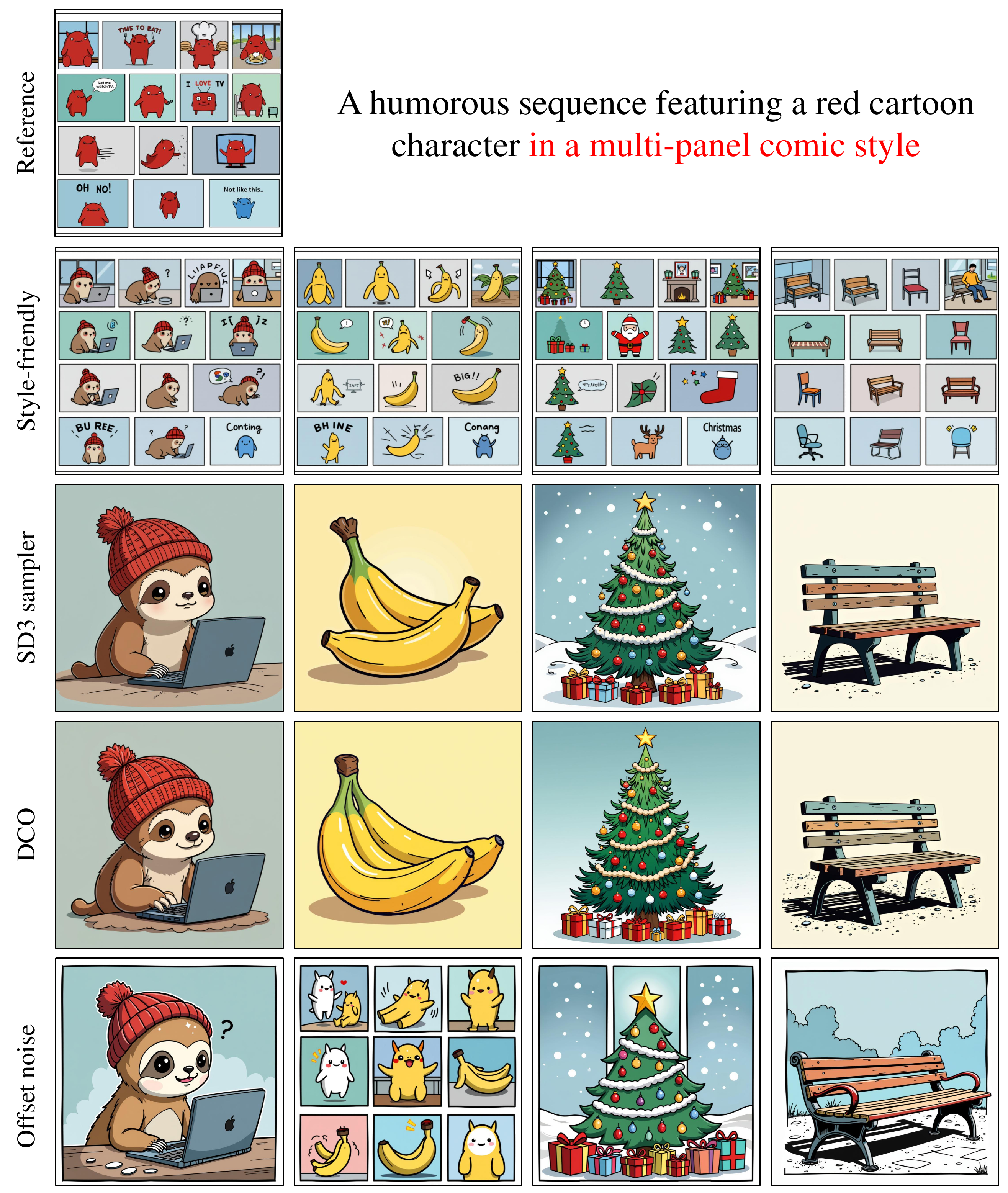}
  \caption{\textbf{Additional qualitative comparison.} Our Style-friendly approach successfully captures complex multi-panel styles, generating images that closely resemble the reference. The prompts used are ``A fluffy baby sloth with a knitted hat trying to figure out a laptop, close up in \{style prompt\} style", ``A banana in \{style prompt\} style", ``A Christmas tree in \{style prompt\} style", and ``A bench in \{style prompt\} style".}
  \label{fig:template_comparison}
\end{figure*}

\begin{figure*}
    \centering
    \includegraphics[width=\linewidth]{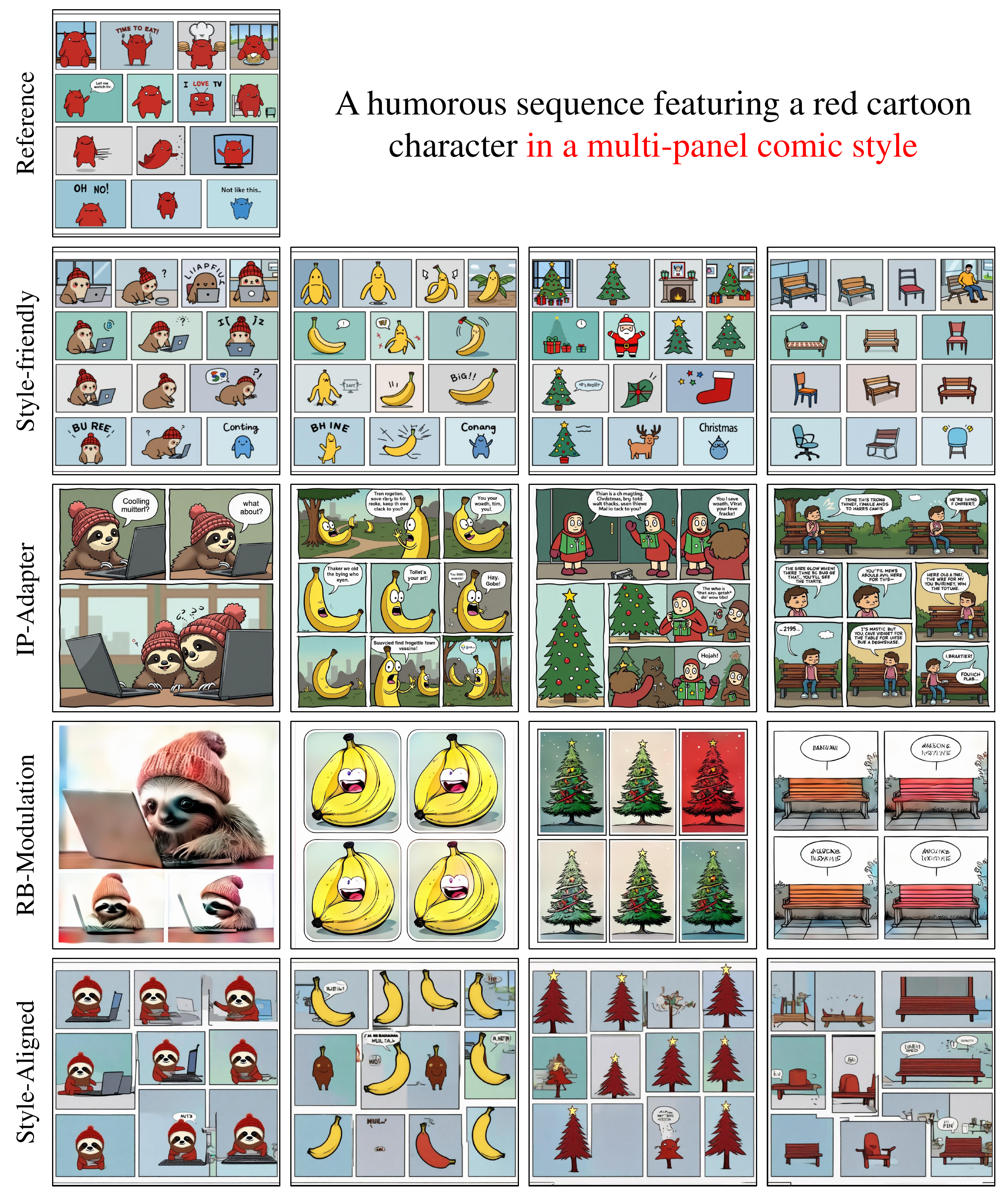}
  \caption{\textbf{Additional qualitative comparison.} Our method effectively captures the multi-panel style, whereas zero-shot methods generate images with different structures or introduce artifacts.}
  \label{fig:template_comparison_zeroshot}
\end{figure*}

\paragraph{Additional Comparison.}
We further demonstrate the effectiveness of a Style-friendly SNR sampler in learning complex style templates, such as multi-panel images. As shown in \cref{fig:template_comparison}, our method captures the given multi-panel style, generating images that closely resemble the reference. In contrast, previous fine-tuning approaches, SD3 sampler~\cite{sd3} and DCO~\cite{dco}, fail to learn the multi-panel concept, producing images without the panel structure. The offset noise~\cite{offset} method attempts to reflect the style but still generates images with a single panel or fewer panels than the reference. Zero-shot approaches including IP-Adapter~\cite{ipadapter}, RB-Modulation~\cite{rb-modulation}, and Style-Aligned~\cite{stylealign} also attempt to generate multi-panel images but often produce outputs with structures different from the reference, as shown in \cref{fig:template_comparison_zeroshot}. This highlights the capability of our method to handle challenging styles that other approaches struggle with.

\begin{figure*}
    \centering
    \includegraphics[width=0.9\linewidth]{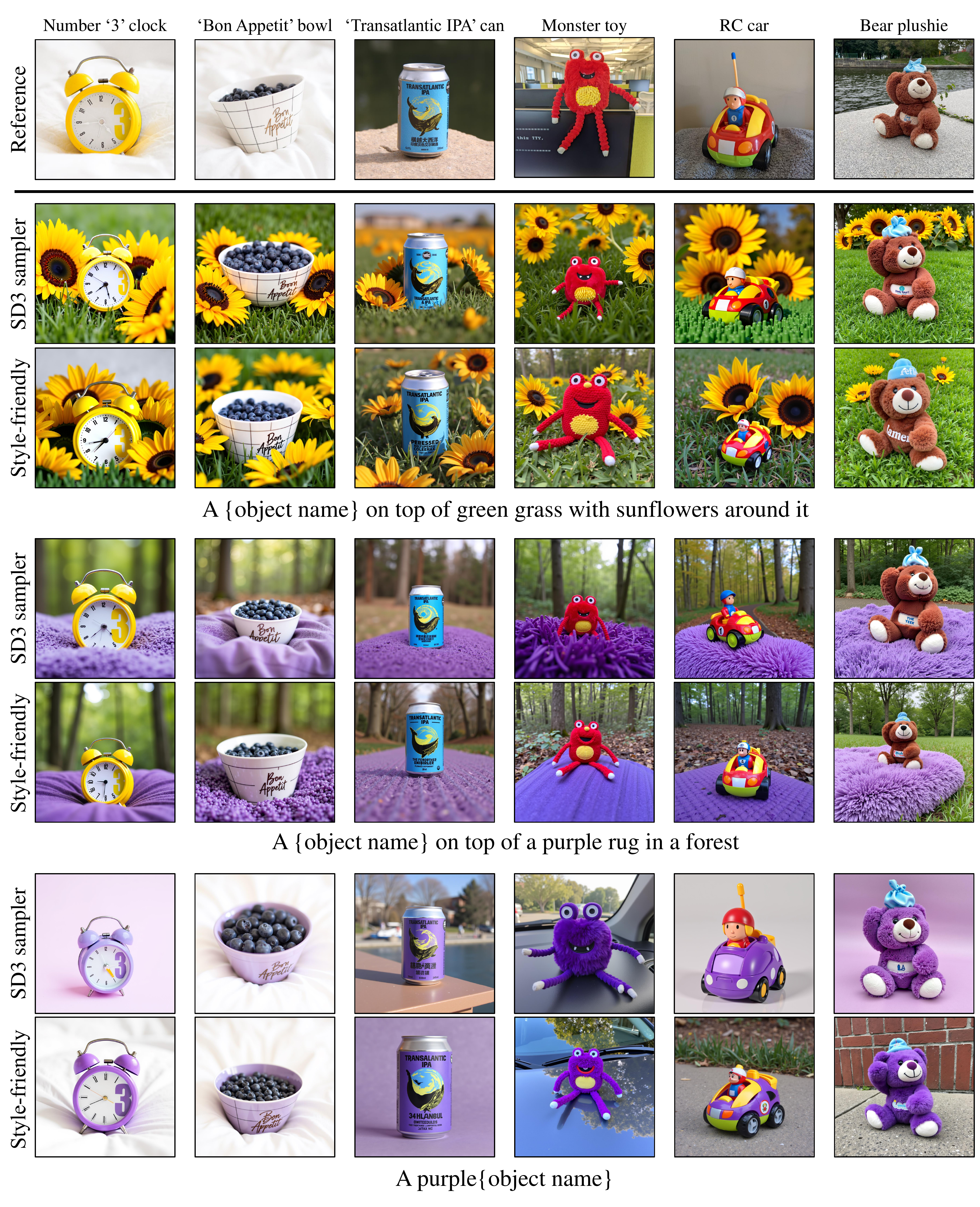}
  \caption{\textbf{Object fine-tuning comparison.} We compare our Style-friendly SNR sampler and the standard sampler on object-driven fine-tuning. Both approaches generate similar overall results, though our Style-friendly sampler occasionally misses minor details, such as the small teeth of the monster toy. Nevertheless, the Style-friendly sampler reliably captures the object's overall shape, color, and key details such as text and patterns on the bowl and can. The object names are written at the top of the reference images.}
  \label{fig:mu_object}
\end{figure*}

\paragraph{Object fine-tuning using Style-friendly SNR sampler.}
We further evaluate whether our proposed Style-friendly SNR sampler—designed specifically for style learning—affects object-driven generation performance. As shown in \cref{fig:mu_object}, our Style-friendly SNR sampler and the original SNR sampler produce qualitatively similar results when fine-tuning on object references. Our approach successfully captures critical details, including shapes, colors, and prominent text or patterns (e.g., on the bowl and can). However, it occasionally omits subtle features, such as the small teeth of the monster toy. These findings support our hypothesis that distinct approaches are required for object-centric and style-driven fine-tuning; our Style-friendly sampler, while slightly suboptimal for object-centric fine-tuning, excels in capturing nuanced style characteristics.

\begin{figure*}[t]
  \centering
   \includegraphics[width=1.0\linewidth]{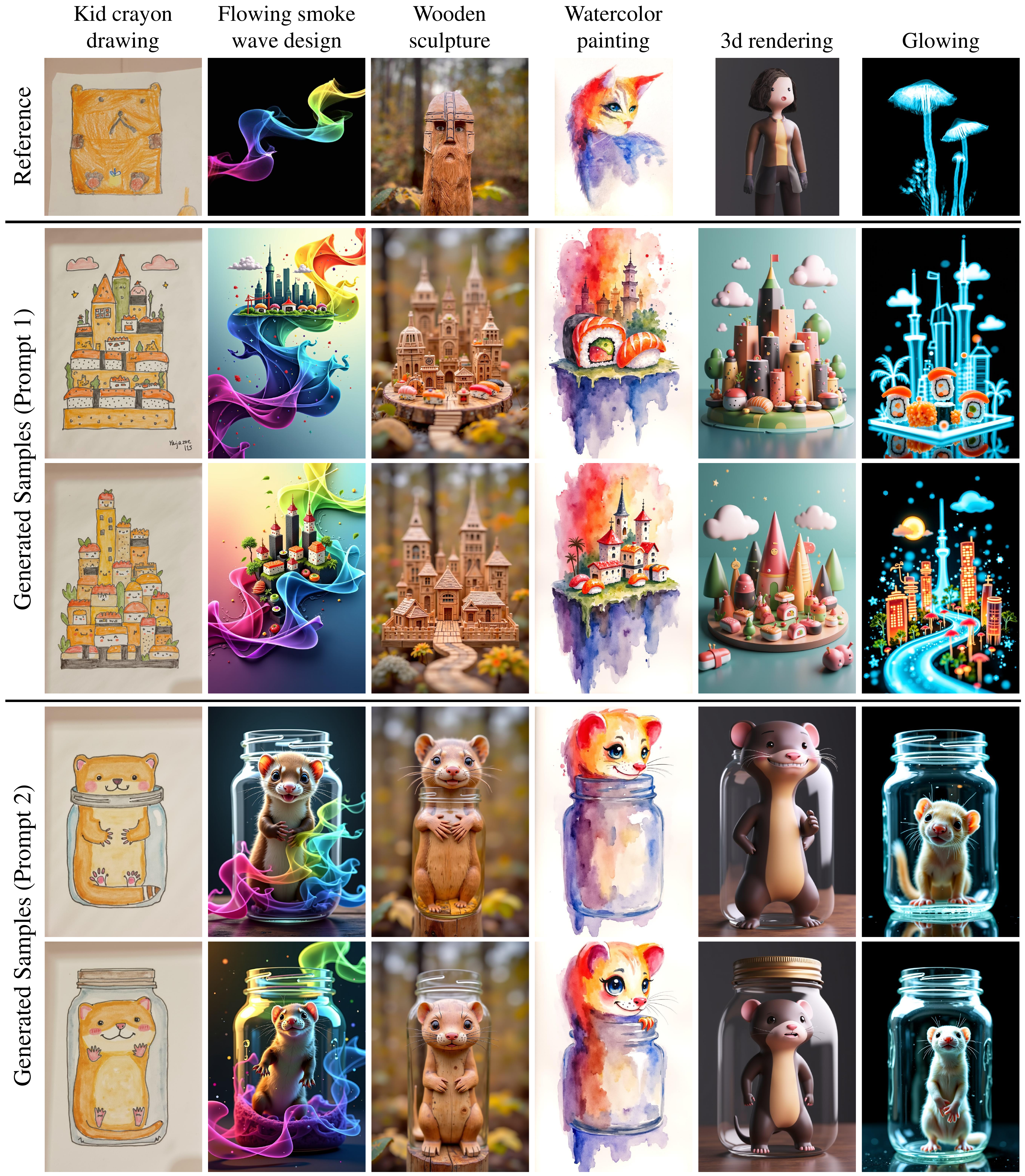}
   \caption{\textbf{Additional samples.} Each row shows images generated with the same random seed at a resolution of 1216×832, using the prompts ``a cute city made of sushi in \{style prompt\} style" and ``mischievous ferret with a playful grin squeezes itself into a large glass jar, in \{style prompt\} style".}
   \label{fig:rectangle}
\end{figure*}

\begin{figure*}[t]
  \centering
   \includegraphics[width=1.0\linewidth]{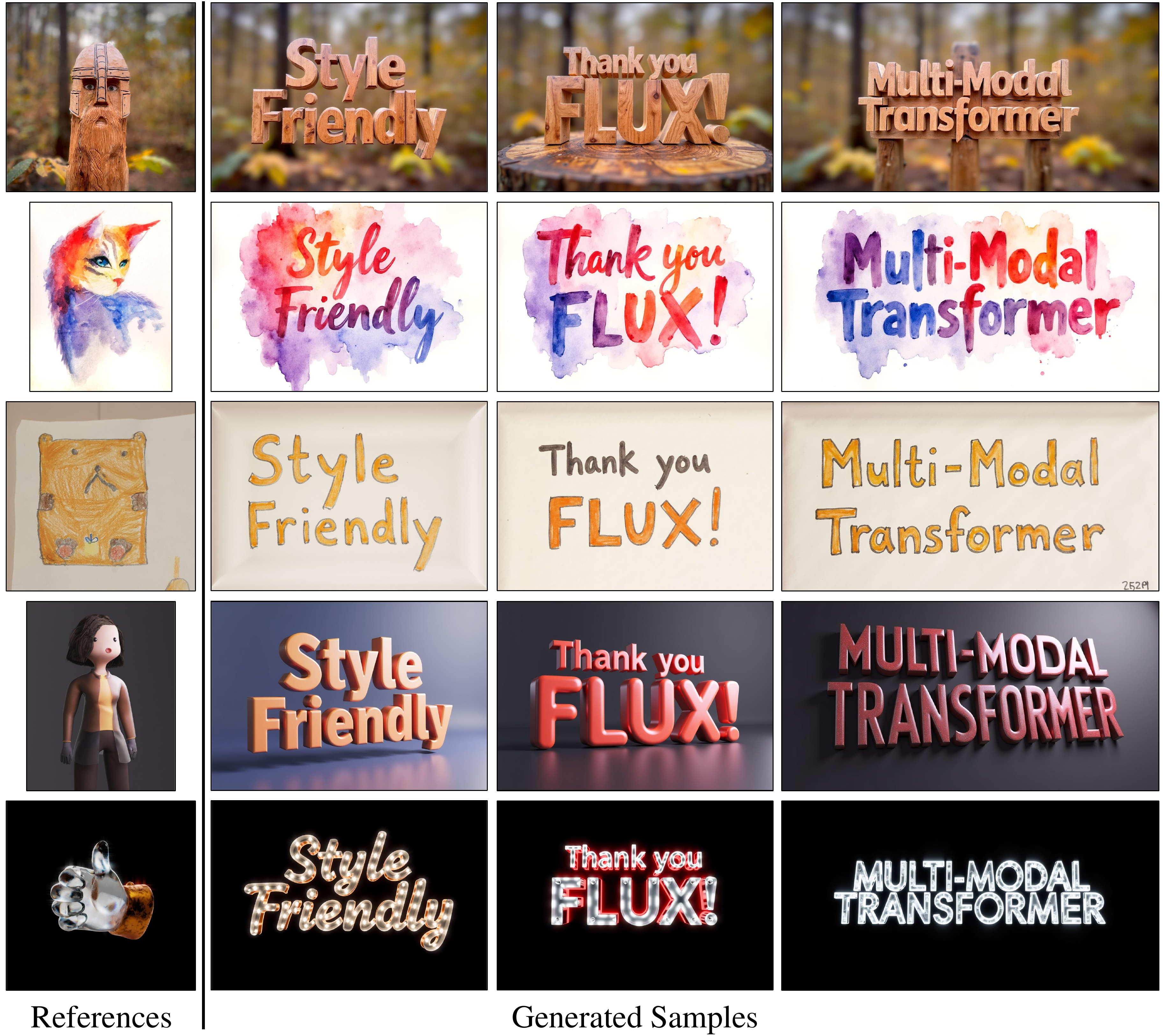}
   \caption{\textbf{Typography.} The first column shows reference images. The second and third columns display samples generated at a resolution of 832×1216, and the fourth column presents samples at 704×1408 resolution. The prompts used are ``the words that says `\{letters\}' are written in English, in \{style prompt\} style", where `\{letters\}' represents the words synthesized in the samples.}
   \label{fig:typography}
\end{figure*}

\paragraph{Additional Samples.}
We present additional samples using the FLUX-dev~\cite{flux1-dev} to demonstrate the versatility of our method. \cref{fig:rectangle} shows that even when fine-tuned on square reference images, our model can generate images with different aspect ratios while maintaining the reference style. For each prompt, we show results from two different random seeds to illustrate diversity across various aspect ratios. \cref{fig:typography} provides additional typography samples in different aspect ratios, exhibiting our capability to produce stylized textual content.

\section{Limitations and Discussions}
\begin{figure*}
  \centering
   \includegraphics[width=1.0\linewidth]{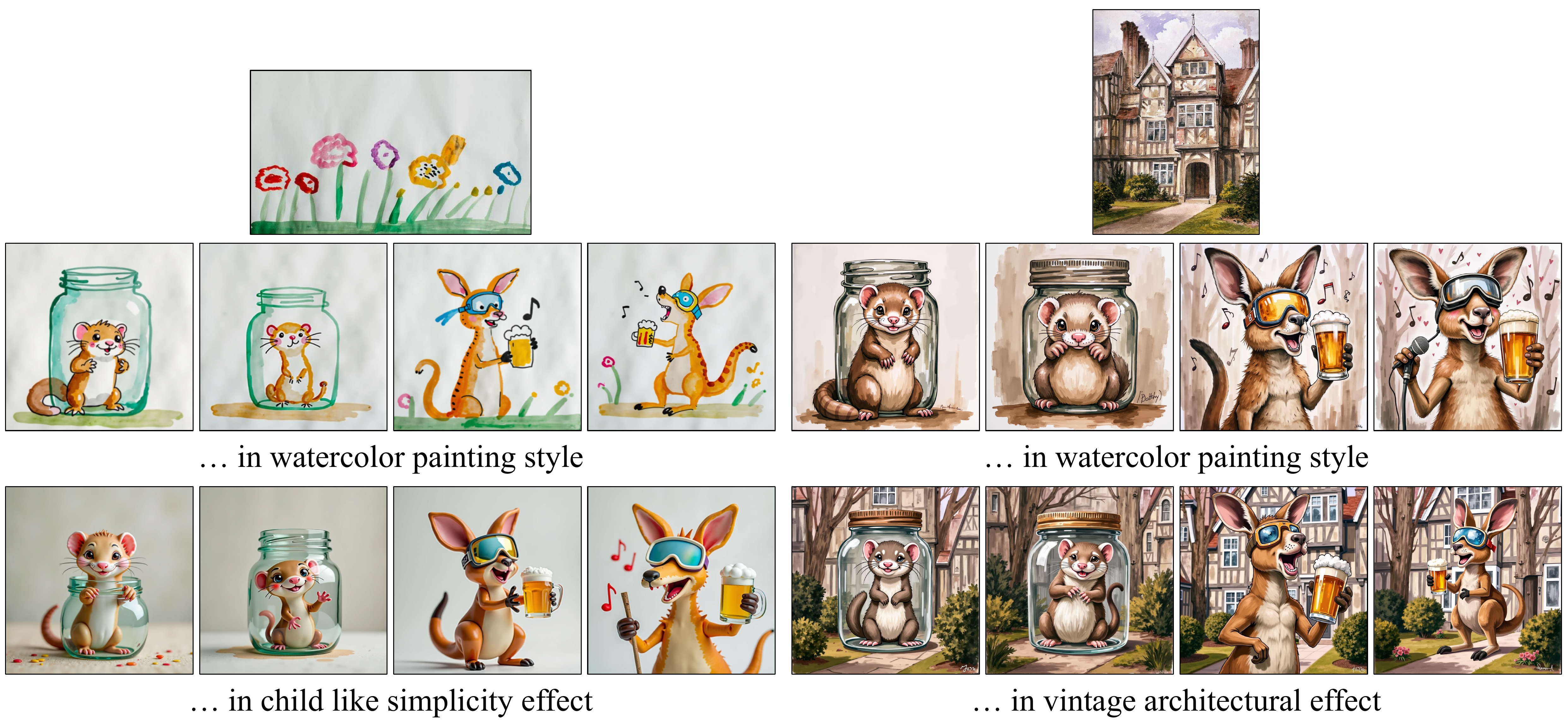}
   \caption{\textbf{Effect of Style Prompt Design.} The first row shows images generated using style prompts from the StyleDrop paper~\cite{styledrop} during both fine-tuning and generation. The second row shows images generated using a different style prompt during both fine-tuning and generation. Each column is generated using the same random seed. This demonstrates how varying the style prompt can lead to different stylistic elements being emphasized in the generated images.}
   \label{fig:prompt_engineering}
   \vspace{35em}
\end{figure*}

\paragraph{Style Prompt Design.}
As shown in \cref{fig:prompt_engineering},  using a different style prompt during fine-tuning can lead to emphasizing different stylistic features, such as child-like elements or background architectures (second row) instead of watercolor painting elements (first row), which may not align with the user's focus. Users should be mindful that variations in the style prompt can lead to different results. Nevertheless, our approach demonstrates effective style learning for style prompts given by the users.

\paragraph{Computational Cost.}
While fine-tuning diffusion models remains the most promising approach for achieving style alignment, it involves significant computational costs. Fine-tuning for a new style typically requires around 300 fine-tuning steps, and due to the iterative nature of diffusion models, generating a single image during inference can take several seconds. We anticipate that future work will explore applying our Style-friendly SNR sampler during the training of zero-shot models~\cite{ipadapter} or integrating it with models that offer faster inference speeds, such as Consistency Models~\cite{consistency} or Adversarial Diffusion Distillation models~\cite{add}. These developments could reduce both training and inference times, making style-driven generation more accessible and efficient.

\section{Broader Impact}
Our Style-friendly SNR sampler makes diffusion models successful in fine-tuning various style references. This advancement allows diffusion models to function effectively as digital art previewers, benefiting artists and non-expert users by simplifying the creative process.
However, we note that it is important to be careful of copyright when using reference images for fine-tuning. Practitioners should ensure they have permissions to use reference images.

\begin{comment}
\section{Rationale}
\label{sec:rationale}
% 
Having the supplementary compiled together with the main paper means that:
% 
\begin{itemize}
\item The supplementary can back-reference sections of the main paper, for example, we can refer to \cref{sec:intro};
\item The main paper can forward reference sub-sections within the supplementary explicitly (e.g. referring to a particular experiment); 
\item When submitted to arXiv, the supplementary will already included at the end of the paper.
\end{itemize}
% 
To split the supplementary pages from the main paper, you can use \href{https://support.apple.com/en-ca/guide/preview/prvw11793/mac#:~:text=Delete%20a%20page%20from%20a,or%20choose%20Edit%20%3E%20Delete).}{Preview (on macOS)}, \href{https://www.adobe.com/acrobat/how-to/delete-pages-from-pdf.html#:~:text=Choose%20%E2%80%9CTools%E2%80%9D%20%3E%20%E2%80%9COrganize,or%20pages%20from%20the%20file.}{Adobe Acrobat} (on all OSs), as well as \href{https://superuser.com/questions/517986/is-it-possible-to-delete-some-pages-of-a-pdf-document}{command line tools}.
\end{comment}

\end{document}